\documentclass[OTHER]{cta-author}

\usepackage{hyperref}
\usepackage{graphicx} 
\usepackage{subcaption} 
\usepackage[utf8]{inputenc}

\begin{document}

\supertitle{Brief Paper}

\title{Tracking Phenological Status and Ecological Interactions in a Hawaiian Cloud Forest Understory using Low-Cost Camera Traps and Visual Foundation Models}

\author{
	\au{Anirudh Potlapally $^{1,*}$}
	\au{Luke Meyers $^{2,*}$}
        \au{Yuyan Chen $^{3}$}
        \au{Mike Long $^{4}$}
        \au{Tanya Berger-Wolf $^1$}
        \au{Hari Subramoni $^{1}$}
        \au{Remi Megret $^{2}$}
	\au{Daniel Rubenstein $^5$}
}

\address{
	\add{1}{Computer Science, The Ohio State University, Columbus, Ohio, USA}
	\add{2}{Computer Science, The University of Puerto Rico Rio Piedras, San Juan, Puerto Rico}
        \add{3}{Computer Science, McGill University, Montreal, Quebec, Canada}
        \add{4}{Battele Ecology, Neon Domain 20, Hilo, Hawaii}
	\add{5}{Ecology, Princeton University, Princeton, New Jersey, USA}
	\add{*}{Both authors contributed equally to this work.}
}



\begin{abstract}
	 Plant phenology, the study of cyclical events such as leafing out, flowering, or fruiting, has wide ecological impacts but is broadly understudied, especially in the tropics. Image analysis has greatly enhanced remote phenological monitoring, yet capturing phenology at the individual level remains challenging. In this project, we deployed low-cost, animal-triggered camera traps at the Pu‘u Maka‘ala Natural Area Reserve in Hawaii to simultaneously document shifts in plant phenology and flora-faunal interactions. Using a combination of foundation vision models and traditional computer vision methods, we measure phenological trends from images comparable to on-the-ground observations without relying on supervised learning techniques. These temporally fine-grained phenology measurements from camera-trap images uncover trends that coarser traditional sampling fails to detect. When combined with detailed visitation data detected from images, these trends can begin to elucidate drivers of both plant phenology and animal ecology. 
\end{abstract}

\maketitle



\section{Introduction}
\label{sec:intro}

Phenology, the study of the recurring seasonal events in plants and animals, such as flowering, fruiting, or migration, is central to understanding ecological processes. These timings influence species interactions, population dynamics, and broader ecosystem functions, and are often sensitive indicators of environmental change \cite{piao2019plant}. While temperate ecosystems are relatively better understood \cite{https://doi.org/10.1111/j.1469-8137.2011.03803.x, https://doi.org/10.1111/nph.14255}, tropical ecosystems remain underrepresented in long-term phenological datasets despite their ecological importance and sensitivity to climate change \cite{10.1093/biosci/biz063, https://doi.org/10.1890/ES12-00299.1}. The National Ecological Observatory Network (NEON) collects standardized phenology data across a network of sites using biweekly in-person surveys and pheno-cam imagery \cite{keller2008continental, elmendorf2016plant}. However, these approaches are labor-intensive and limited in spatial and temporal resolution, often missing rapid phenophase transitions and subtle plant–animal interactions \cite{tang2016emerging}.
In this study, we use motion-triggered camera traps to monitor phenology at the scale of individual plants in a complex tropical forest understory. While these cameras are primarily intended to capture animal movement, they often incidentally record phenological changes that occur alongside faunal activity or other sources of motion in the environment, such as strong wind. Compared to NEON's fixed-interval sampling, camera traps provide higher temporal resolution and capture interactions that occur between scheduled field observations. Yet, the resulting datasets are large in volume and temporally and spatially irregular, presenting significant analytical challenges. 

To overcome these challenges, we adopt a hybrid computer vision approach that combines pre-trained foundation models with simple, task-specific image processing techniques. By leveraging these powerful models in our workflows, we reduce reliance on supervised training, allowing traditional computer vision-based processing to complement and denoise model predictions. This allows us to perform tasks such as object detection and trait estimation without requiring large expert annotated datasets or custom-trained models, significantly reducing the computational and human costs typically associated with supervised learning pipelines.

This study presents a scalable and low-cost framework for monitoring fine-scale phenological change and associated animal interactions using motion-triggered camera traps and automated computer vision workflows. Our main contributions are as follows: 
\begin{enumerate}
  \item Design and deploy a dual-mount camera-trap system capable of capturing both above- and below-canopy plant structures and faunal activity in a tropical forest understory.
  \item Develop an integrated analysis pipeline that combines foundation vision models with traditional image processing techniques to extract phenological and ecological interaction signals from irregular, unstructured image data without annotation and supervised training. 
  \item Validate our method's temporal sensitivity and practical utility by comparing it to NEON's biweekly field observations, showing that our system captures rapid transitions and interactions that conventional methods may overlook.
\end{enumerate}
Our results show that low-cost camera traps, combined with off-the-shelf vision models, can effectively capture fine-scale phenological patterns and plant–animal interactions in complex tropical understory environments, offering a practical approach that could potentially be extended to other ecosystems.

\begin{figure}[th]
\centering
\includegraphics[width=.45\textwidth]{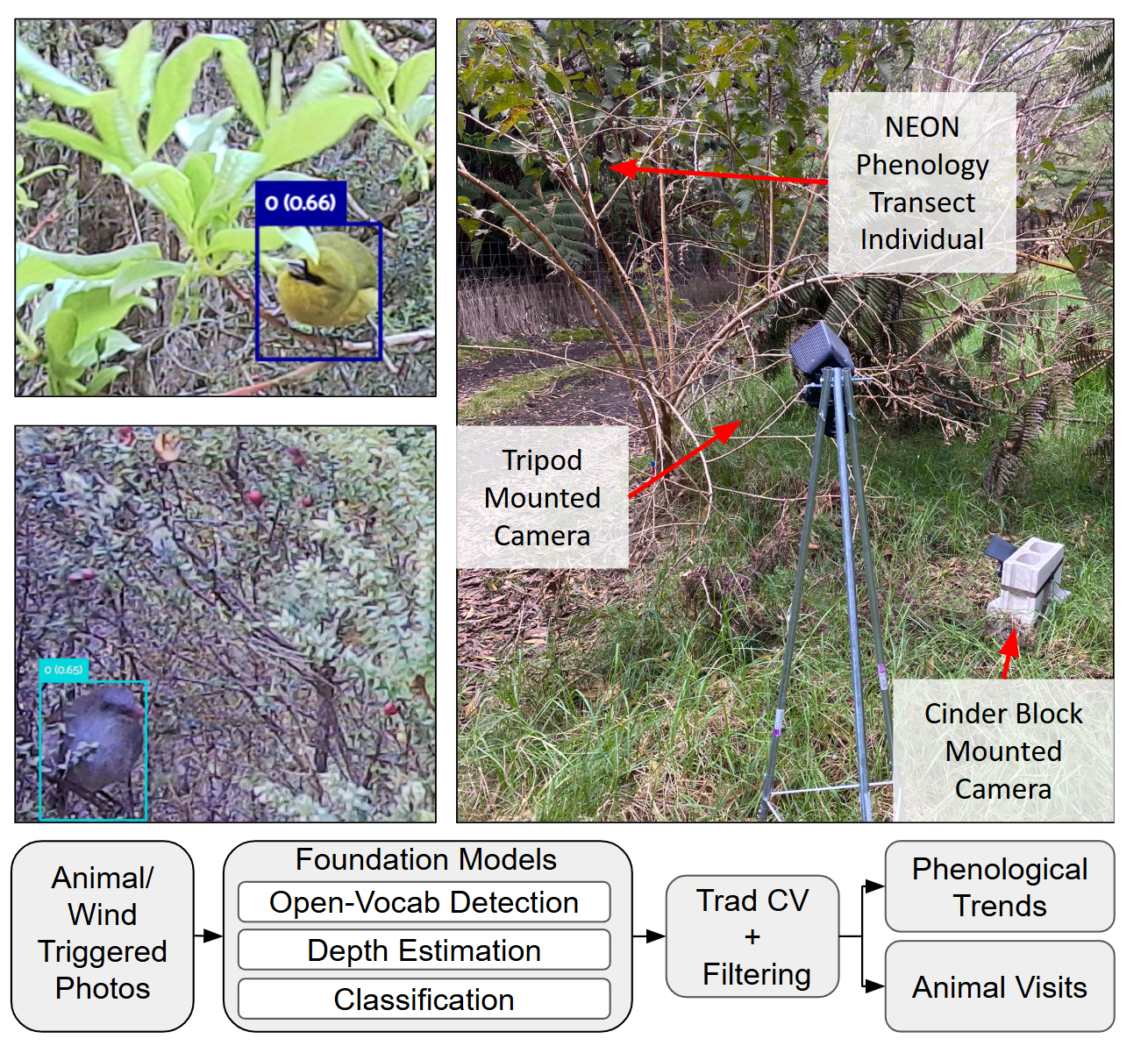}
\caption{
Top Left: Crop showing an 'amakihi (\textit{Chlorodrepanis virens}) visiting an ohelo flower. 
Bottom Left: An 'omao (\textit{Myadestes obscurus}) consuming a pukiawe berry. 
Right: Camera‐trap setup—tripod and cinderblock‐mounted—targeting a \textit{Rubus hawaiiensis} individual along NEON’s phenology transect.
Below: Mini-overview of general workflow. 
}
\label{fig:birds}
\end{figure}

\section{Background}
\label{sec:background}

Phenology is an integrative science that studies the timing of seasonally occurring biological events, which can have important ecological impacts such as influencing species distributions, demographic trends, and ecosystem functions like nutrient cycling \cite{morisette2009tracking, piao2019plant, chuine2010does, elmendorf2016plant}. Plant phenological transitions, such as leafing out, flowering, or fruiting, are influenced by natural cues such as daylight, freezing, or other climatic variables, many of which are expected to shift significantly in the coming decades \cite{inouye2022climate, ovaskainen2013community}. Shifts in phenological timing can alter species interactions \cite{https://doi.org/10.1002/ecy.3846}, disrupt pollination networks, and affect biogeochemical cycles \cite{doi:10.1139/X08-104}. In many temperate systems, long‐term records have revealed clear advances in spring leaf‐out and flowering in response to warming trends \cite{https://doi.org/10.1111/j.1469-8137.2011.03803.x, https://doi.org/10.1111/nph.14255}. However, tropical ecosystems remain comparatively understudied, despite their high biodiversity and risk for high impacts from climate change \cite{10.1093/biosci/biz063, https://doi.org/10.1890/ES12-00299.1}. 

The National Ecological Observatory Network (NEON) operates a continent-wide infrastructure of instrumented sites designed to collect standardized ecological and environmental data across diverse biomes to monitor continent-scale environmental change \cite{keller2008continental}. At each domain, NEON primarily relies on detailed phenological observations gathered through weekly or biweekly in-person field surveys conducted along small transects embedded within the larger landscape. Additionally, NEON supplements ground-based surveys with automated "pheno-cam" imagery to further enhance the monitoring of phenological changes across broader spatial scales \cite{elmendorf2016plant, richardson2018tracking}. Even with these data, phenology remains challenging to study due to its inherent variability across spatial and temporal scales \cite{tang2016emerging, PARK2021709}. Traditional pixel-based indices used for analysis of pheno-cam imagery work well at the ecosystem level, but may miss more detailed or nuanced interactions \cite{richardson2018tracking}. At the individual plant level, detailed in-person observations are labor-intensive and limited in spatial coverage and temporal resolution \cite{tang2016emerging}. As a result, they often miss rapid phenological transitions or subtle interactions between plants and animal visitors, such as pollinators and frugivores, making it challenging to capture phenological dynamics at scales relevant to individual organisms or species interactions \cite{tang2016emerging}. These limitations hinder efforts to link observed patterns to environmental drivers and reduce our ability to generalize across sites or taxa.

Camera‐trap techniques \cite{tanwar2021camera, 10.3389/fevo.2021.617996}, long used in wildlife ecology to document animal presence and behavior, offer a complementary approach for phenological study. Efficient motion-triggered cameras capture images when movement is detected—often from animals, but occasionally from strong wind or other disturbances—resulting in datasets that are concentrated around biologically active periods. Recently, these methods have been employed to study more detailed plant-animal interactions, such as fruit dispersal \cite{villalva2024frugivory} or pollination \cite{krauss2018effectiveness}. Even though they do not directly record plant changes, they frequently capture phenological transitions coinciding with animal visitation \cite{monteza2022arboreal}. While established methods exist for detecting or classifying animals and other potential triggers from camera trap imagery\cite{beery2019efficient}, previous work connecting plant change and animal activity has only used manual phenophase annotation \cite{hofmeester2020using} or the same low precision techniques applied to larger scale imagery \cite{sun2021simultaneous, de2022camera} Compared to NEON's biweekly field observations, camera trap based event-driven sampling provides greater temporal precision and sensitivity to ecological interactions. However, working with camera-trap datasets presents a number of practical challenges: image quality is often inconsistent, the timing of captures is irregular, and the sheer volume of data can make manual review slow and resource-intensive. These limitations point to the need for automated approaches to extract relevant ecological information to analyze plant-animal interactions at scale. 

Computer vision has emerged as a powerful method for automating image analysis and has a long history of being used across multiple domains like healthcare \cite{JAVAID2024792}, geospatial analysis \cite{HARIPAVAN2025100371, rs17020179}, agriculture \cite{9144058}, and ecology \cite{10.1145/3626186, DHANYA2022211}.Form  Early techniques depended on hand-crafted feature extraction, where visual characteristics were manually defined and used as input for machine learning models \cite{Chandra2021, 6975001, electronics12051199} performing tasks such as classification or segmentation. This approach shifted with the advent of Deep Convolutional Neural Networks (CNNs) \cite{NIPS2012_c399862d, He_2016_CVPR, https://doi.org/10.1155/2018/7068349}, which allowed models to learn rich feature representations directly from raw image data. While highly effective, these models require large, annotated datasets for training, thus costly in data-scarce ecological settings \cite{TalaeiKhoei2023, doi:10.1139/er-2018-0034}. While transfer learning \cite{Hosna2022, gupta2022deep, zhao2024comparison} has enabed some pre-trained models to learn new tasks, they ultimately rely on substantial labeled data in the new domain. These methods therefor remain task-specific and continue to depend heavily on domain-relevant data to perform effectively \cite{10.1145/3582688}.

Recently, computer vision and deep learning research have moved from application-specific models toward large, generalizable "foundation" models that can be adapted to a range of vision and language tasks. Transformer-based architectures—originally developed for natural language \cite{NIPS2017_3f5ee243}—were adapted to visual data \cite{dosovitskiy2021an}, giving rise to multi-modal models such as CLIP \cite{pmlr-v139-radford21a} or ALIGN \cite{DBLP:conf/icml/JiaYXCPPLSLD21}, which learn joint image–text representations from paired data. These multimodal models leverage the semantic information within text to develop a better understanding of visual input, and have been used to create domain-specific variants such as BioCLIP \cite{stevens2024bioclip} and MedCLIP \cite{wang2022medclip}. Subsequent extensions, including models like Flamingo \cite{10.5555/3600270.3601993} and GPT-4 Vision \cite{2023arXiv230308774O}, further integrated these modalities, enabling zero-shot classification \cite{pmlr-v139-radford21a, 10377550}, object detection \cite{10.1007/978-3-031-72970-6_3, 10657519, 10.5555/3666122.3669313}, segmentation \cite{ravi2025sam}, depth estimation \cite{ke2025marigold, depth-pro}, captioning \cite{wang2022git, 10.5555/3618408.3619222}, retrieval \cite{Suma_2024_ECCV, 10378372}, and even reasoning \cite{li2019visualbert, wang2023image}about scene composition without task-specific retraining. Many of these latest systems rely on agentic calling of multiple tasks or domain-specific foundation models, which perform better than a multipurpose model for all needs \cite{deitke2024molmo, team2023gemini, bai2025qwen2}. This flexibility reduces the need for large, labeled datasets in every new application and simplifies the workflow for disciplines such as ecology, where field-collected imagery and sensor data must be analyzed at scale. 

To address the challenges of scaling fine-grained phenological monitoring, we used foundation models in combination with traditional computer vision techniques to develop a streamlined, generalizable analysis pipeline. Using imagery from low-cost, motion-triggered camera traps deployed around a diverse set of understory species at a NEON site, our goal was to extend phenological observations to in situ individuals under natural environmental conditions and faunal interactions—moving beyond the constraints of time-lapse systems typically used in semi-controlled settings \cite{crimmins2008monitoring}. This approach is designed to support and complement labor-intensive field-based methods by reducing the reliance on manually collected data, which is costly, difficult to scale, and often too sparse to capture fast or subtle transitions.

Given the large volume of unlabelled images and the limited feasibility of expert annotation, we employ foundation models capable of performing complex vision tasks out-of-the-box—such as object detection \cite{10.1007/978-3-031-72970-6_3, 10.5555/3666122.3669313, beery2019efficient}, species classification \cite{stevens2024bioclip}, and monocular depth estimation \cite{depth-pro} for foreground segmentation—without the need for supervised training. These outputs are combined with lightweight, task-specific computer vision techniques: we use color space transformations to identify red-channel signals associated with fruiting \cite{10136612}, estimate greenness \cite{quevedo2025pixel} to monitor canopy development or flowering, and isolate regions of interest based on phenological traits. This hybrid approach reduces the need for labeled data while enabling consistent, high-throughput processing of large and variable image datasets. By standardizing the workflow across plant species and imaging contexts, our method provides a scalable and low-cost strategy for monitoring phenological change and ecological interactions across domains.

\section{Data Acquisition at NEON Domain 20 Pu'u Maka'ala Natural Area Reserve}

\subsection{Data}

We deployed low-cost trail cameras around ten individuals of NEON-monitored understory species to capture phenological changes and faunal activity in and around the focal plants. The cameras were mounted both on tripods at chest height to frame the plant canopy and on cinder blocks at ground level to monitor activity at the base. NEON scientists swapped SD cards approximately every 15 days during routine transect visits. The tripod-mounted cameras recorded seasonal changes in flowers and foliage as well as visiting birds and mammals, while the ground-mounted units captured terrestrial fauna moving near the plant base. By using inexpensive, easily assembled supports alongside affordable scouting cameras, we maximized our sample size and increased spatial resolution to test generalizable methods on more taxa. However, this approach is not without limitations. The low-cost cameras, primarily marketed as hunting aids, are triggered by passive infrared motion sensors and lack a timed capture function. Their low-resolution sensors produce significant image noise, and the mounts are often disturbed during SD card replacements. Additionally, because phenology is monitored remotely, drastic changes in plant structure or growth over time—and the emergence of new branches—can shift or obstruct the field of view.

The combination of factors makes our dataset unique and challenging for monitoring phenology, as it has neither temporal nor positional consistency. In total, 12,484 images were collected between January 27 and April 3, spanning roughly two months at the site. The data follow a classic long-tailed distribution, with a few cameras producing significantly more triggers than others. In general, the top seven cameras were associated with focal species situated in more open areas of the forest, where movement caused by wind—combined with close proximity to the sensor—frequently triggered the camera. In this work, we present an analysis of a subset of species that completed significant phenological shifts at the site during the data collection window.

\subsection{Site} 
We conducted our preliminary study at NEON Domain 20, located within the Pu'u Maka'ala Natural Area Reserve on Hawai'i Island. The reserve, which spans 18,730 acres, was established in 1981 to protect culturally significant and biodiverse native rainforest ecosystems. Its management efforts include invasive species control, habitat restoration, and the protection of watershed and cultural values \cite{dnr_2013}. The site was incorporated into the NEON network as Domain 20, and the ongoing collection of in-person ecological observations at this location is critical for validating remote sensing–based monitoring systems such as the one we propose. Within the NEON network, Pu'u Maka'ala stands out as a highly diverse tropical cloud forest. Globally, tropical phenology remains poorly understood \cite{piao2019plant}, and even within the moist tropical forests of Mauna Loa, the phenological drivers of dominant canopy species were only begun to be investigated in 2022 \cite{pau2020climatic}. Insights from our work contribute to filling this gap by focusing on understudied species within this unique observatory site.
In addition to its ecological and cultural significance, the reserve provides critical habitat for several threatened native bird species, offering refuge from mosquito-borne diseases and serving as a stronghold for native plant communities \cite{10.1650/CONDOR-18-25.1}. Many of the rare plant species in the reserve are known to be bird-pollinated \cite{10.2984/73.2.1}, presenting opportunities to observe both phenological patterns and biotic interactions for rare and important taxa.
\section{Approaches and Methods}
\begin{figure}[t]
    \centering

    \begin{subfigure}[b]{0.49\columnwidth}
        \centering
        \includegraphics[width=\linewidth]{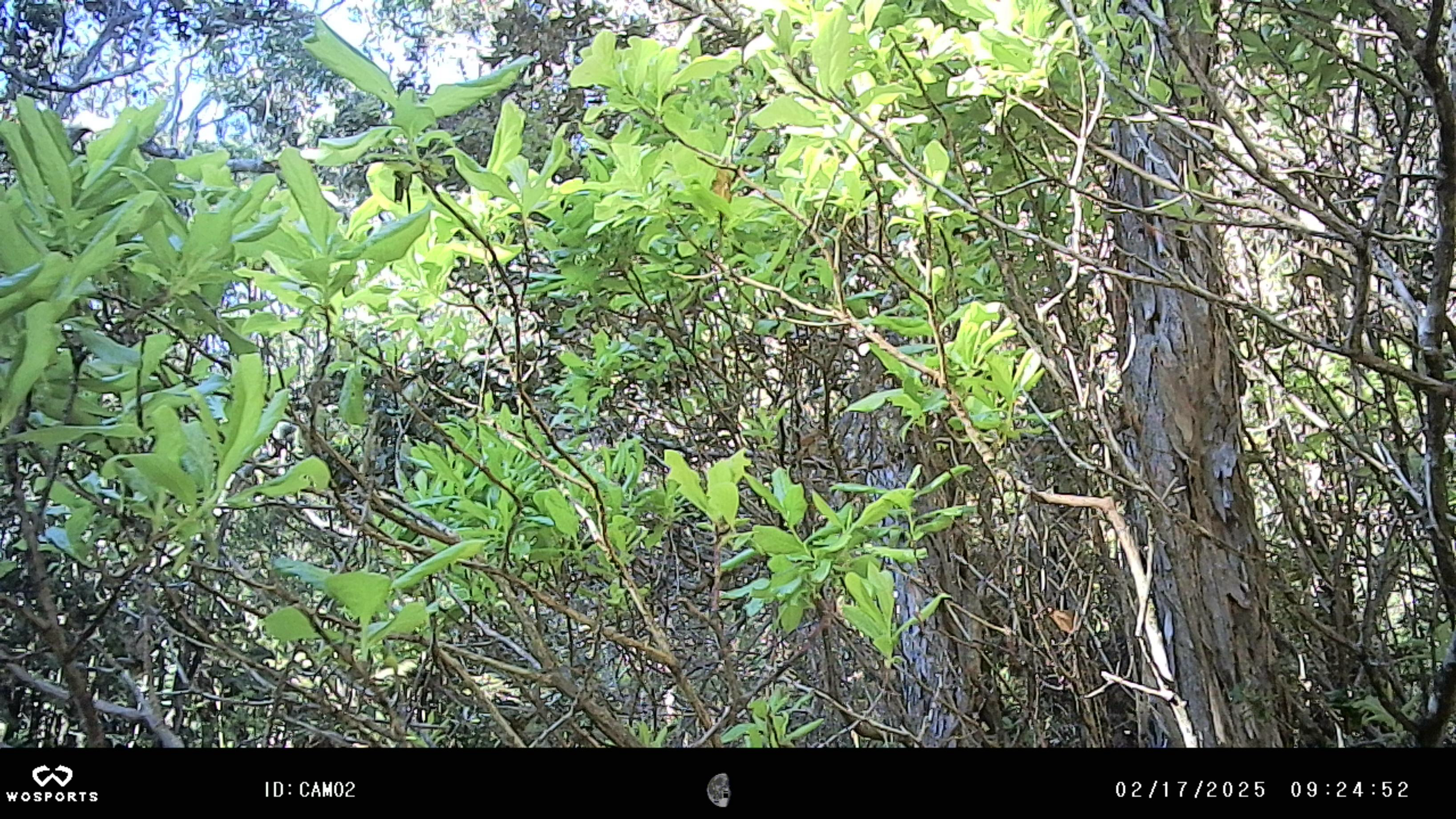}
        \caption{Raw Camera Trap Image}
        \label{fig:greenness_pipeline:1}
    \end{subfigure}
    \hfill
    \begin{subfigure}[b]{0.49\columnwidth}
        \centering
        \includegraphics[width=\linewidth]{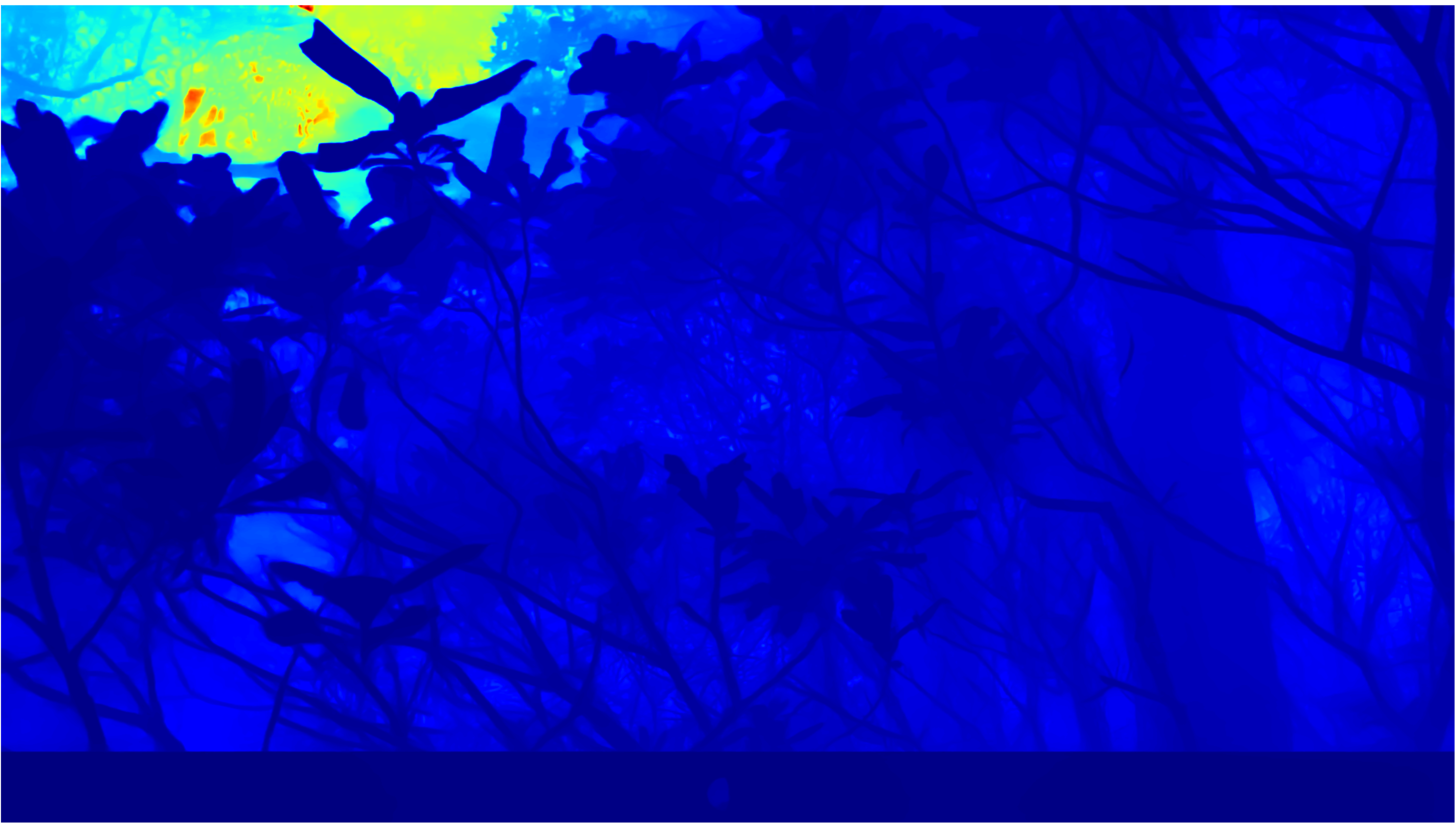}
        \caption{Depth Map From DepthPro}
        \label{fig:greenness_pipeline:2}
    \end{subfigure}

    \vspace{1ex}

    \begin{subfigure}[b]{0.49\columnwidth}
        \centering
        \includegraphics[width=\linewidth]{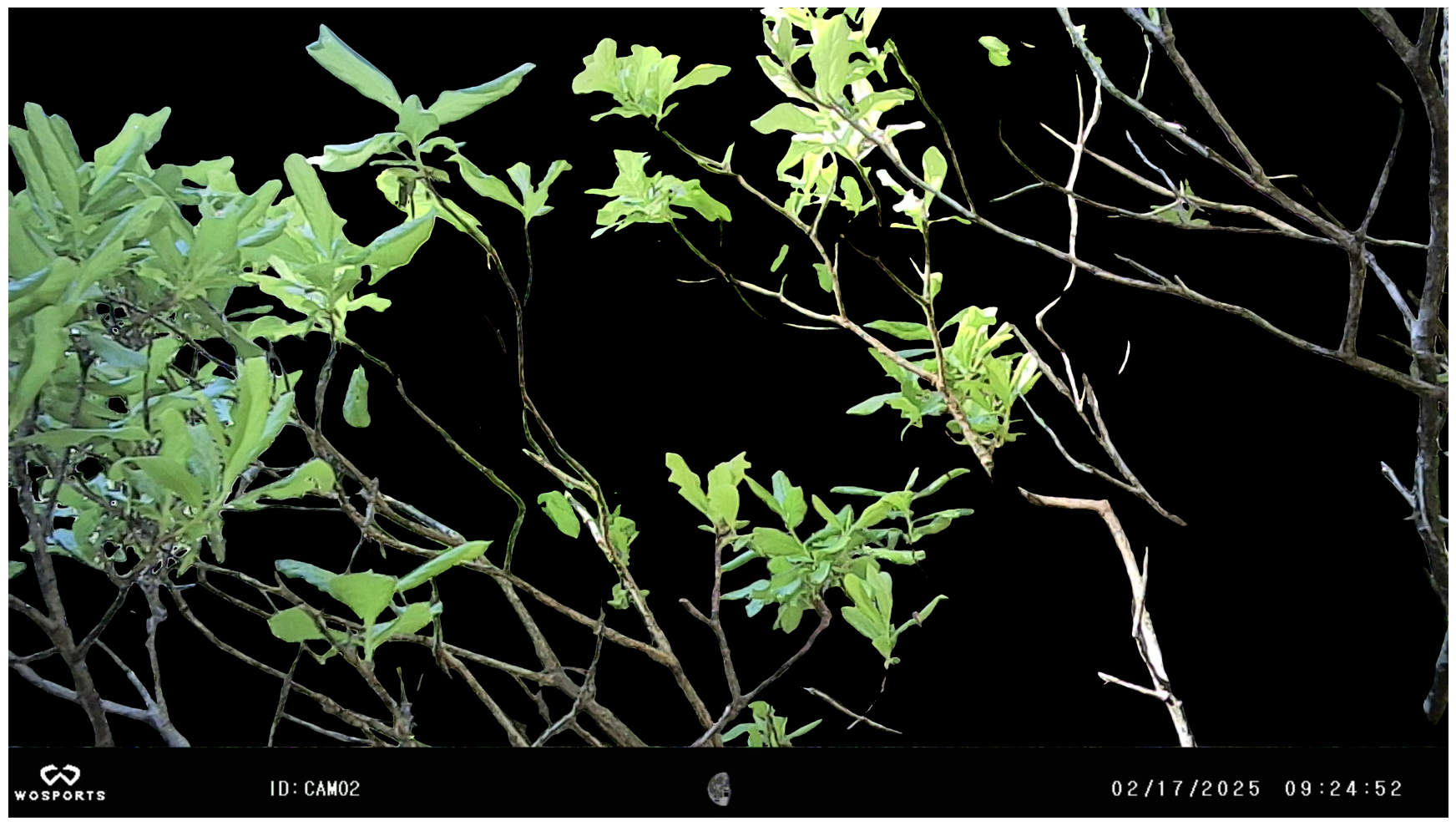}
        \caption{Foregrounded Extraction}
        \label{fig:greenness_pipeline:3}
    \end{subfigure}
    \hfill
    \begin{subfigure}[b]{0.49\columnwidth}
        \centering
        \includegraphics[width=\linewidth]{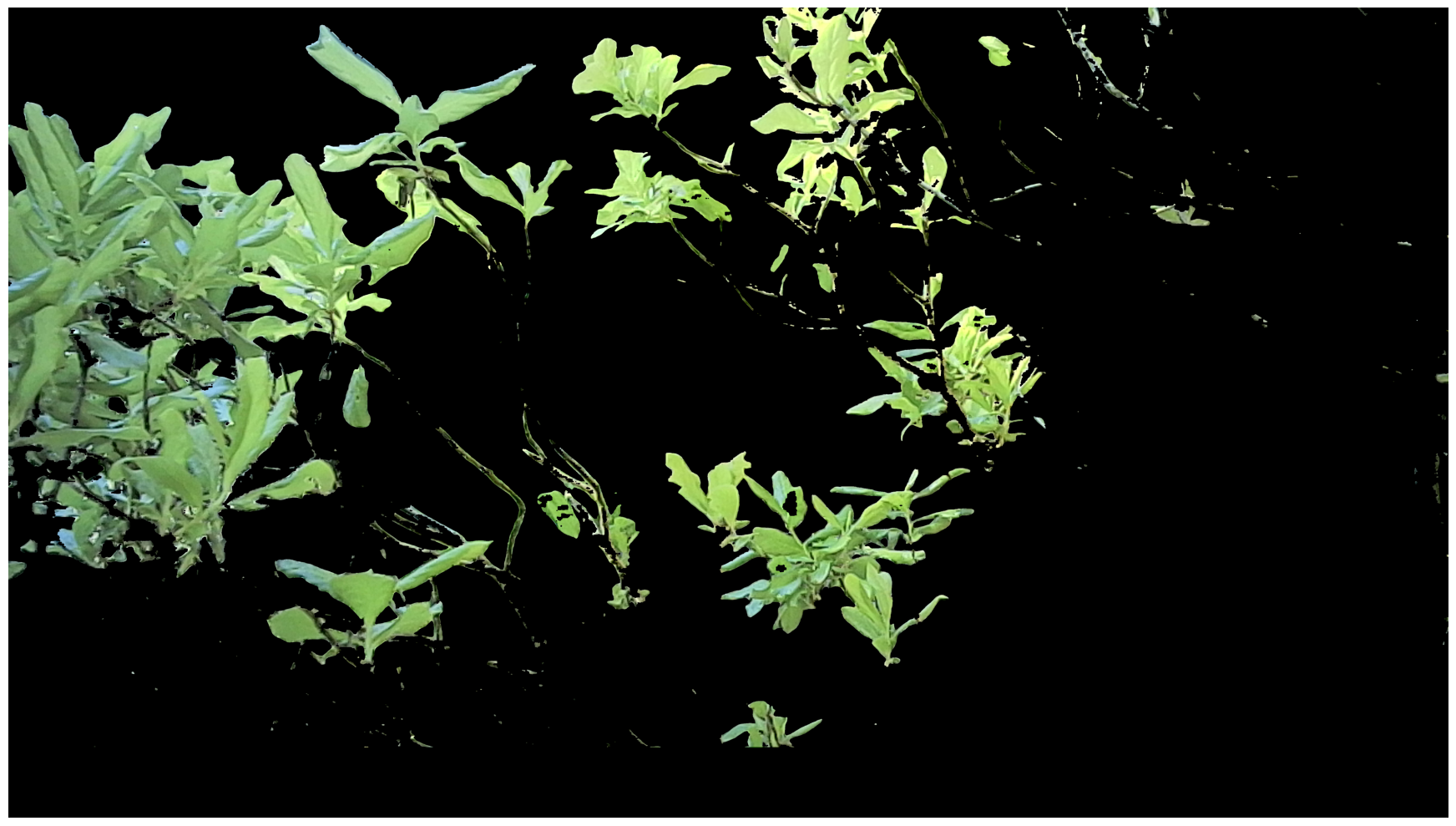}
        \caption{Thresholding Green Pixels}
        \label{fig:greenness_pipeline:4}
    \end{subfigure}

    \caption{\textbf{Canopy Leaf Cover Estimation} (a) Raw image of the focal plant against a complex background, making individual leaf counts difficult. (b) Depth map from DepthPro; applying a 2 m threshold isolates the plant foreground. (c) Extracted foreground showing only the focal plant with background removed. (d) Segmented leaves obtained by converting the foreground to LAB space and isolating green-chromaticity pixels.}
    \label{fig:greenness_pipeline}
\end{figure}
\subsection{Leafing in Individual 'Ōhelo, \textit{Vaccinium calycinum}}

We examined leaf cover change in individuals of forest 'ōhelo, \textit{Vaccinium calycinum}. This species is unique among Hawaiian vacciniums for its deciduousness, but the character is weak and highly variable between individuals \cite{becker2024evolution}. For our studied individuals, while a clear change in canopy leaf cover is apparent in the image sequences, quantifying this change is non-trivial, as separating the foliage of the target species from the complex background of the Hawaiian cloud forest is challenging. Although supervised methods have success in foliage segmentation, to reduce our dependence on extensive annotation, we explored zero-shot monocular depth estimation using DepthPro \cite{depth-pro}. We first temporally aligned each image sequence to track the same plant individual consistently. Each image sequence was first temporally aligned to ensure consistent tracking of the same plant individual. For every frame, a depth map was generated using DepthPro, and an empirically determined threshold was applied to isolate the foreground (i.e., the focal plant) from the background. Although the threshold value was held constant for each individual, it was adjusted across different plants to account for variation in lighting conditions. 

The resulting foreground masks were converted into LAB color space—separating luminance (L) from chromaticity (a and b) channels—and used the 'a' channel to identify green pixels. A greenness score was computed by calculating the percentage of green pixels within each region of interest. Repeating this workflow across the full image set yielded a time series of greenness scores for each plant. Figure~\ref{fig:greenness_pipeline} illustrates the stages of processing involved in estimating canopy greenness. Once greenness values were obtained, spurious measurements were removed by applying DBSCAN \cite{ester1996density}, which flagged outliers as noise points. A low-degree polynomial regression model was then fit to the cleaned data to capture the underlying phenological trend, and the coefficient of determination ($R^2$) was computed to quantify the model's goodness of fit. A similar polynomial model was fit to NEON's manually recorded canopy leaf cover estimates collected during bi-weekly field visits.

\subsection{Fruiting in Pūkiawe, \textit{Leptecophylla tameiameiae}}
Berry production was tracked for a highlight pūkiawe, \textit{Leptecophylla tameiameiae}, specimen. Detecting berries in images proved challenging for pretrained deep learning approaches due to their small size; however, their distinctive red coloration led us to explore a dual color-space segmentation approach. Each image captured by the camera was processed in both HSV (using dual red-hue thresholds to capture both bright and dark reds) and LAB color spaces (by thresholding the a-channel). After cleaning each mask using morphological opening and closing to reduce noise, we extracted contours exceeding a minimum area threshold (50 square pixels) and computed their centroids. HSV and LAB detections were then matched by pairing contours whose centroids lay within 50 pixels of one another; matched pairs were merged into single bounding boxes and counted as estimated berries per frame.

The raw berry counts were compiled into a chronological sequence to model temporal patterns. DBSCAN was applied to identify and remove spurious detections, retaining only inlier points for further analysis. A second-order polynomial regression model was fit to these inliers, and the coefficient of determination ($R^2$) was computed to assess model fit. As with the leaf cover estimation experiment, model's berry count trends were compared with the fruit-count observations recorded by NEON scientists during their routine manual inspections.

\subsection{Avian Visitor Detection}

To filter images and identify faunal visitors, we tested several generalizable detection models with classification-based filtering. MegaDetector \cite{beery2019efficient} is a general‐purpose object detection model for camera‐trap images—classifying animals, humans, and vehicles—and serves as the standard tool for removing “empty” frames. We also compared two SOTA zero-shot object detection models: Grounded DINO \cite{10.1007/978-3-031-72970-6_3} and OWLv2 \cite{10.5555/3666122.3669313}. As vision-language models, these systems require both an image and a text prompt. For our use case, each image was paired with the prompt "bird" and processed to generate bounding boxes. Detections from all models were subsequently passed to BioCLIP \cite{stevens2024bioclip} for filtering of false positives, and in the case of MegaDetector, non-bird animals. BioCLIP predicts taxonomic classification from kingdom to species with strong performance, especially on rare or unseen taxa. We used the built-in TreeOfLife classifier in the pybioclip package to categorize animal detections at the class level. Based on these predictions, we sorted detections into "bird" (classified as \textit{Aves}) and "not bird" (any other class) pseudo-labels. Finally, we tested the addition of some low-level filtering of detections based on expectations of the system, such as bird movement and behavior. Detections that shared Intersection-over-Union (IoU) more than .75 for 5 consecutive detections were considered automatic false positives. In all comparisons, models were run with the same threshold, 0.2 for detection confidence. Detections were evaluated using an IoU greater than .1, as tight bounding boxes were not annotated for our data, and multi-individual frames were scarce, with no overlap between individual predictions. 

After detection, bird visitors to pūkiawe and 'ōhelo were manually annotated to species for further ecological analysis. The pipeline was evaluated on 4,625 images from four cameras collected during the ~2-month preliminary data collection period. For future refinement, we also extracted BioCLIP's internal 512-dimensional image embeddings for these evaluation images, enabling additional exploration of clustering and classification strategies. Once bird visits were detected, they were stitched temporally into visits if the time between camera captures was less than 15 seconds. The resulting unique visit counts by species per day were analyzed alongside  NEON phenological observations and automated berry counts for pūkiawe. Additionally, bird visits were examined across flowering for two individuals of 'ōhelo, \textit{Vaccinium calycinum}. We fit a low-degree polynomial regression model to capture the underlying phenological trends and compute the coefficient of determination (R²) to quantify goodness of fit.
\section{Results and Discussion}

\begin{figure}[t]
\centering
\begin{subfigure}[b]{0.45\textwidth}
    \includegraphics[width=\linewidth]{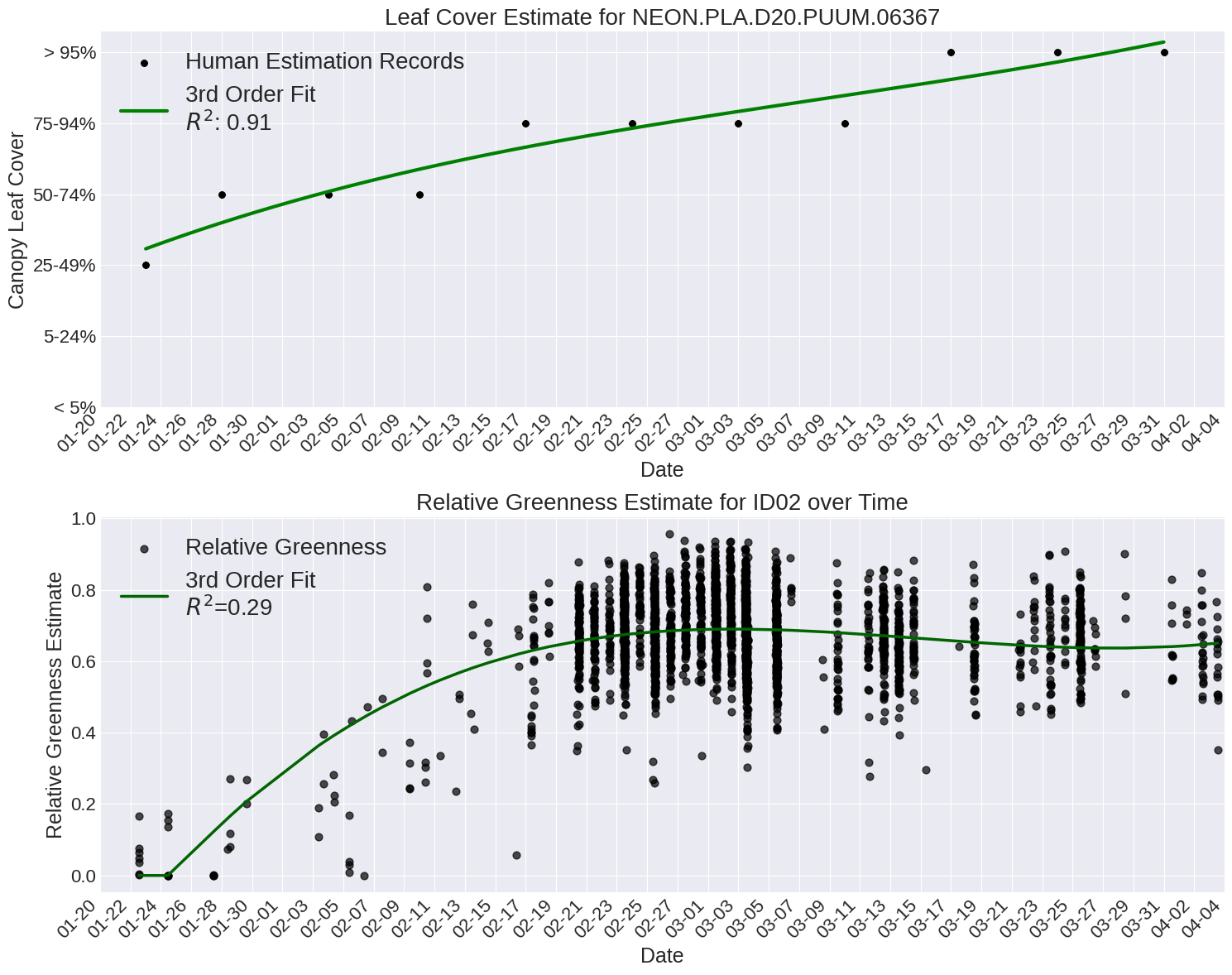}
    \label{fig:img1}
\end{subfigure}
\begin{subfigure}[b]{0.45\textwidth}
    \includegraphics[width=\linewidth]{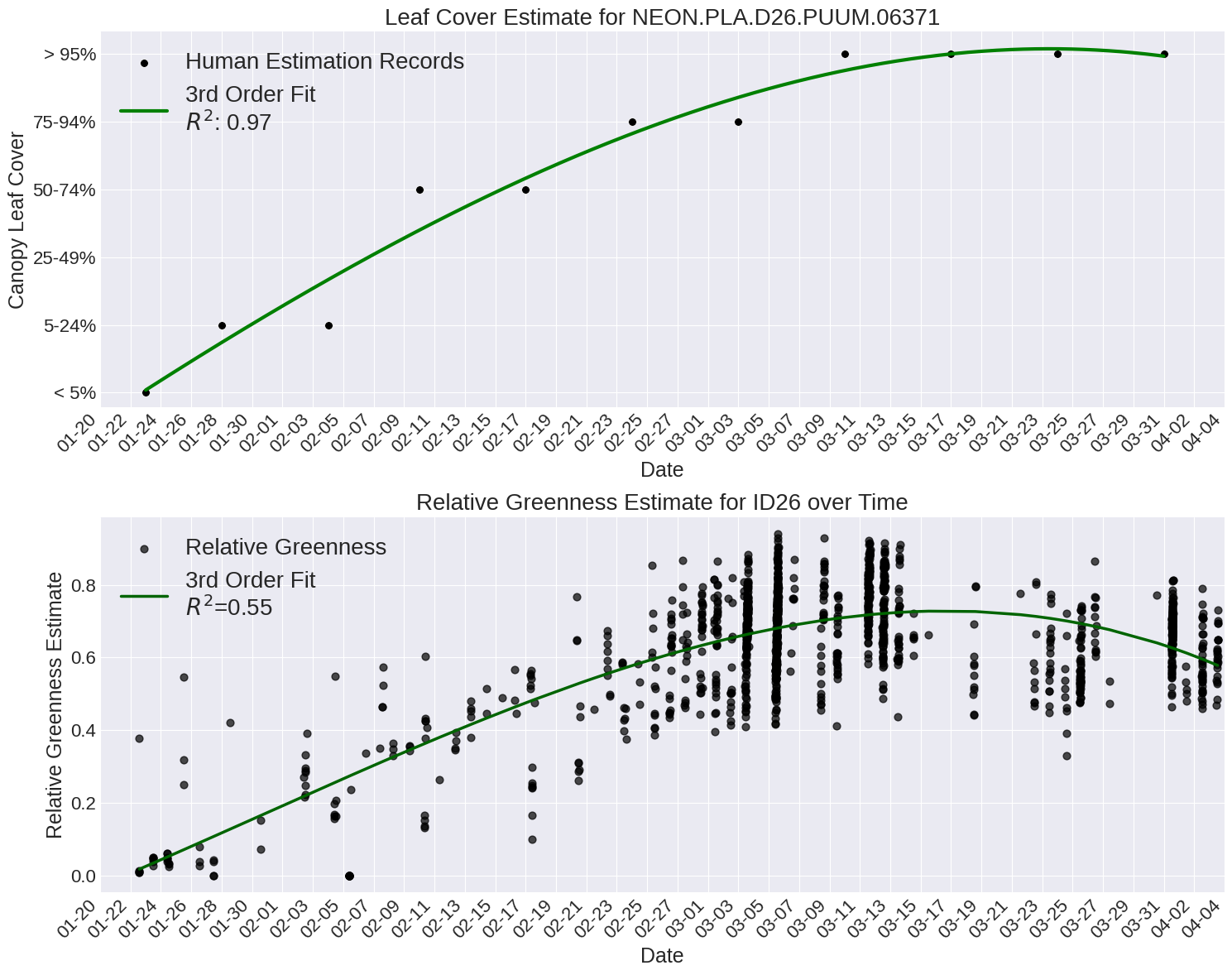}
    \label{fig:img2}
\end{subfigure}
\caption{Change in greenness of Ohelo (Vaccinium calycinum) canopy.
Top: NEON manual canopy‐cover estimates and our method's greenness estimates for specimen 06367 on the PUUM phenology transect.
Bottom: NEON observations and our greenness estimates for specimen 06371.
The strong correspondence between methods demonstrates that our approach accurately captures the trends recorded by NEON scientists.
}
\label{fig:side_by_side}
\end{figure}

Applying computer vision and deep learning to quantify greenness in 'ōhelo canopies yielded results that align well with NEON's manual observations. As shown in Figure~\ref{fig:side_by_side}, our leaf-cover estimates closely track NEON's biweekly records, capturing the sharp increase in canopy density from late January to late February and its stabilization in early March ($R^2~\approx~0.91-0.97$). The use of cameras at each plant enabled higher temporal resolution than periodic field surveys, generating a more detailed time series of canopy change.

In contrast, relative-greenness estimates exhibited greater day-to-day variability ($R^2~\approx~0.29-0.55$), likely due to sensitivity to transient factors like lighting and shadows. Although this increased data density introduced greater variability in the raw measurements, fitting a polynomial regression shows an overall trajectory consistent with NEON's observations. These findings demonstrate that vision-based methods can recover key phenological signals and, with proper calibration, offer a scalable path toward individual-level, automated monitoring. Additionally, depth-based foreground extraction allowed us to isolate canopy features without requiring manual annotation, highlighting its value for fine-scale phenology analysis.

\begin{figure}[ht]
\centering
\includegraphics[width=.5\textwidth]{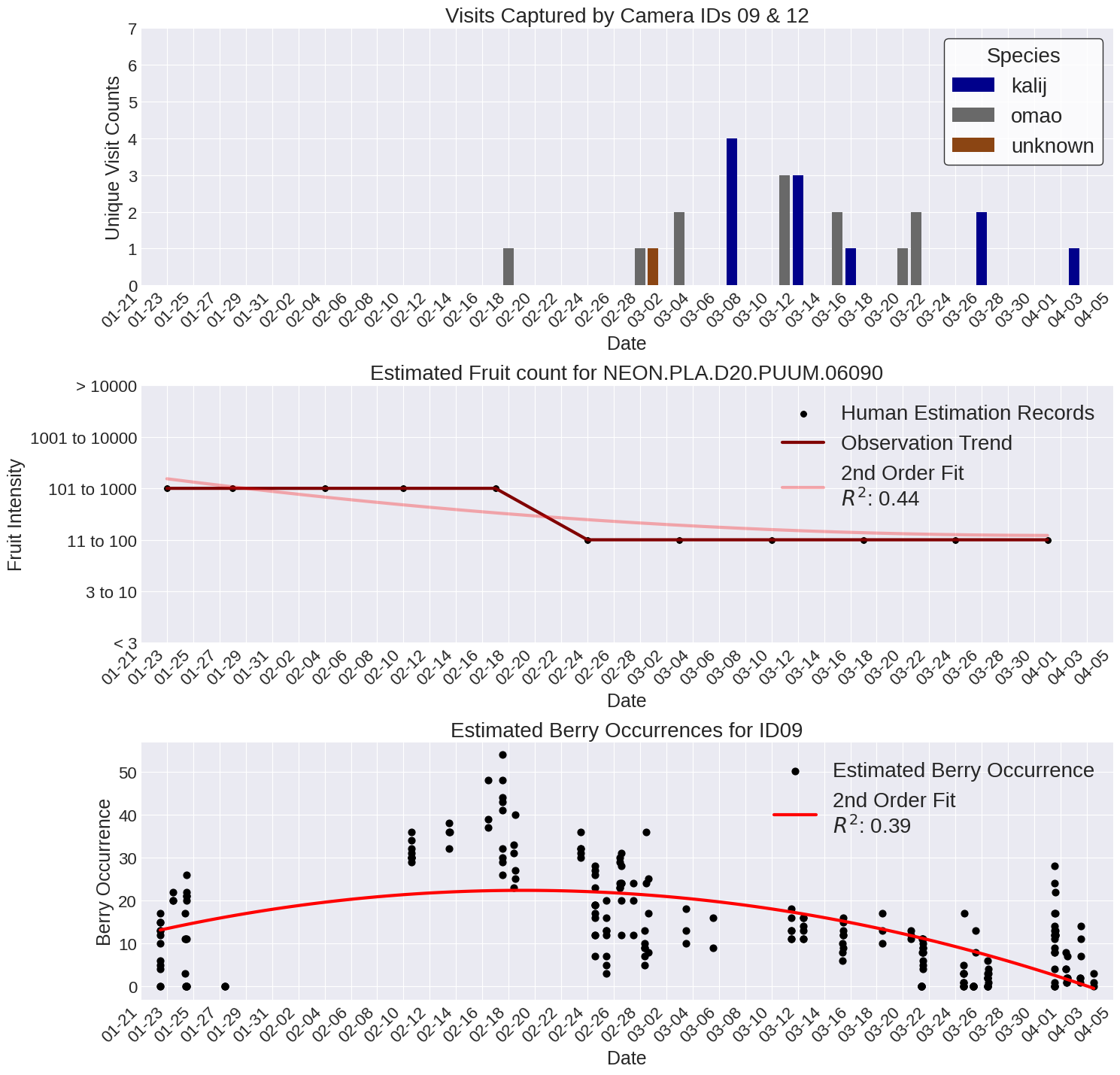}
\caption{\textbf{Avian Visitors and Berry Count Estimates of Pukiawe through Time}. Top: Number of bird visits at specimen 06090 on the Phenology Transect. Middle: Manual berry counts by NEON scientists for specimen 06090. Bottom: Berry counts estimated by our computer vision model.  
}
\label{fig:berries}
\end{figure}

Automated berry‐production estimates for pūkiawe peaked in early February and then declined steadily through March. Figure \ref{fig:berries} shows our raw berry detections showed high day‐to‐day variability, but a second‐order polynomial fit ($R^2 = 0.39$) demonstrates the same inflection point of berry decline ($\approx$ Feb 20) as a step-wise function of NEON estimates. Because our pipeline relies entirely on traditional computer vision steps—color-space segmentation, morphological filtering, contour matching, and clustering—instead of supervised learning, fixed thresholds can be sensitive to shifts in lighting, berry color saturation, and partial occlusion, which amplifies short-term fluctuations in the counts. 

When we overlaid bird‐visit data, however, a clear pattern emerged: omā‘o (\textit{Myadestes obscurus}) visits to the canopy rose as berry abundance leveled off (peaking in early March), suggesting feeding on the ripe fruits, causing the decrease in estimates of relative abundance. Later in the season, ground‐level detections of Kalij pheasant (\textit{Lophura leucomelanos}) increased, suggesting berries had begun to fall off the plant. These results demonstrate that high-frequency camera-based measurements can not only replicate NEON's biweekly phenology snapshots without supervised models, but also uncover fine-scale dynamics of plant–bird interactions that periodic human surveys alone would miss.



\begin{table}[!b]
\processtable{Zero Shot Detection Comparison\label{tab1}}
{\begin{tabular*}{20pc}{@{\extracolsep{\fill}}llll@{}}\toprule
Detection Method         & Precision     & Recall        & F1             \\
\midrule
GD (GroundingDINO) Only            & 0.055         & \textbf{0.855} & 0.103         \\
GD + BC (BioCLIP)                  & 0.342         & 0.805         & 0.480         \\
GD + BC + Filtering                 & 0.389         & 0.805         & 0.524         \\
OWLv2 Only                          & 0.783         & 0.751         & 0.767         \\
OWLv2 + BC                          & \textbf{0.873} & 0.715         & \textbf{0.786} \\
OWLv2 + BC + Filtering              & \textbf{0.873} & 0.715         & \textbf{0.786} \\
MD (MegaDetector) Only             & 0.199         & 0.828         & 0.320         \\
MD + BC                             & 0.366         & 0.760         & 0.494         \\
MD + BC + Filtering                 & 0.421         & 0.756         & 0.540         \\
\botrule
\end{tabular*}}{}
\end{table}

%


\begin{figure}[th]
\centering
\includegraphics[width=.5\textwidth]{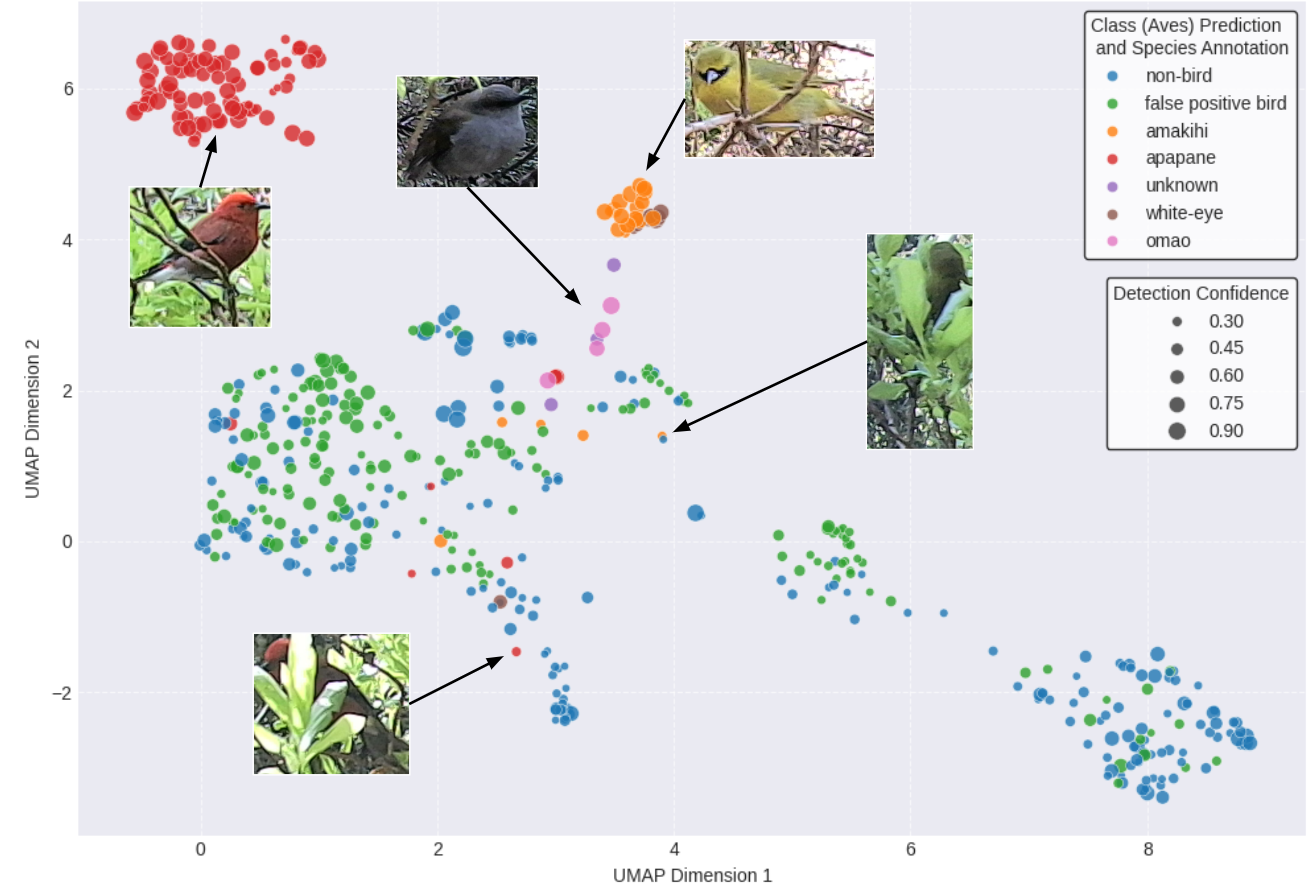}
\caption{UMAP visualization of BioCLIP features of positive detection crops at Camera 02 and 26, non-bird and false positive bird are MegaDetector final predictions, species labels annotated by hand}
\label{fig:birds}
\end{figure} 

Table \ref{tab1} summarizes bird-detection performance on a subset of 4,625 images that together contained 221 confirmed bird visits. MegaDetector's default confidence threshold of 0.20 is tuned for high recall, yet Grounding DINO—run with a lowered threshold—achieved the highest recall (0.855). Much of this gain came from over-prediction, but after BioCLIP re-classification and filtering, Grounding DINO still retained a recall advantage over MegaDetector while being only 0.04 less precise. OWLv2 provided the most balanced output up front, yielding an F1 score of 0.77 and markedly fewer false positives than either alternative. Because OWLv2 was already precise, BioCLIP could improve its precision by only 0.09, whereas the same filtering boosted MegaDetector and Grounding DINO by factors of 1.8× and 6.2×, respectively. Taken together, these results suggest that OWLv2 offers a strong out-of-the-box option, while Grounding DINO benefits most from downstream filtering if maximum recall is required.

We further examine the error in detection and the future potential of species classification in Figure \ref{fig:birds}, which shows a UMAP \cite{mcinnes2018umap} projection of the BioCLIP features per cropped detection output from MegaDetector. Each point represents a predicted image, colored by species (or predicted class) and scaled by the detector's confidence score. Species labels were hand-annotated, and false positives are distinguished by whether BioCLIP classified them as "Aves." The projection reveals well-separated clusters for true positives—especially for 'apapane, 'amakihi, and Japanese white-eye—while 'ōma'o appears closer to a central "zone of confusion," likely because of its darker plumage. The tight clustering of known species suggests that false positives could be reduced by allowing BioCLIP to assign more specific labels or applying a K-nearest-neighbor classifier to BioCLIP features, enabling taxonomic discrimination beyond a simple "bird/not-bird" threshold. 

Qualitative inspection of the bird crops that fall outside UMAP species clusters offers useful clues to understand pipeline error. These outliers contain birds that were detected, but are small, heavily occluded, or blurred by motion. Since BioCLIP embeds and classifies based on cropped detections, the dominant occlusions' features outweigh the bird's signature, and the model assigns a plant label. This behavior underscores the need for object-centered images in models such as BioCLIP, and how detection models may be more robust to difficult samples that may fail in multipurpose classification models. In practice, the value of BioCLIP filtering depends on both the detector's performance and study priorities: if recall matters more than fine-grained taxonomy, OWLv2's unfiltered outputs already deliver the best balance of precision and recall, making further classification less critical.

\begin{figure}[t]
\centering
\includegraphics[width=.50\textwidth]{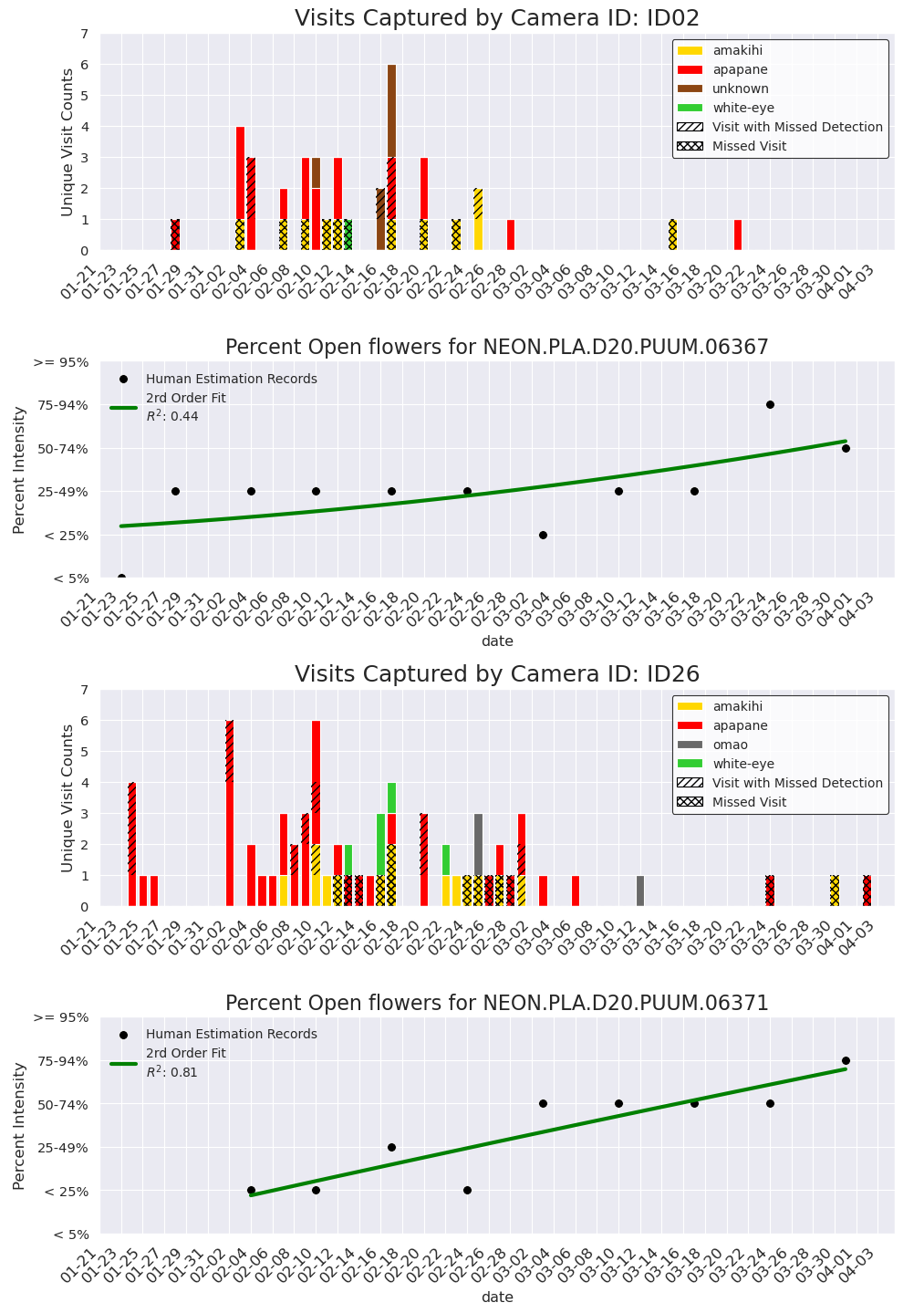}
\caption{ Avian visitors and open-flower counts for Ohelo (Vaccinium calycinum) specimens over time.
Top \& Third: Bird visits at specimens 06367 and 06371 on the Phenology Transect, colored by species. "Visits with missed detections" are visits where OWLv2 missed frames, while "missed visits" failed to be detected entirely.  
Second \& Bottom: Manual open-flower counts by NEON scientists for specimens 06367 and 06371, respectively.}
\label{fig:oheloflowers}
\end{figure}

Using verified bird detections on two 'ōhelo (\textit{Vaccinium calycinum}) individuals, we reconstructed daily pollinator visitation over time. In Figure \ref{fig:oheloflowers}, the top panels plot unique visits per day—colored by species and compared against OWLv2 combined outputs—while the bottom panels show NEON technicians' weekly "percent flowers open" measurements for the same plants. Bird activity peaks just before the midpoint of flower opening. We hypothesize this matches the expected maximum bloom, as older flowers drop and new buds open, percent-open flowers rises until all flowers are in bloom, then declines. Although directly detecting these small green flowers in our images was infeasible without further annotation, the pronounced spikes in bird visits coincided with qualitative estimation from images of peak bloom. Additionally, the clear succession in visitor composition —'apapane arrive first, followed by 'amakihi and Japanese white-eye at the height of flowering — further shows ecological insights that can be gained from these methods.

\section{Conclusions and Future Work}

In this study, we demonstrated that combining off-the-shelf AI models with traditional computer-vision routines can capture key phenological and ecological patterns in situ. By deploying low-cost, motion-triggered cameras, we generated high-frequency time series of canopy leaf cover and berry production that closely matched those of NEON's manual observations, but at a much finer resolution. Concurrently, we tracked wildlife visits to two understory shrubs, linking pollinator and frugivore activity to flowering and fruiting phases. These findings show that image-based workflows—even without supervised training—can produce ecologically meaningful phenological metrics, delivering relevant measurements without the need for dedicated annotation.

Leveraging foundation models without any additional training proved both practical and insightful. DepthPro's zero‐shot monocular depth estimation effectively isolates individual plants, and Grounded DINO and OWLv2—though not specifically designed for camera‐trap imagery—performed on par with MegaDetector for object detection. BioCLIP further enhanced animal classification within our pipeline. At the same time, our berry-detection case study underscores the ongoing value of traditional computer vision: dual-color-space segmentation, morphological filtering, contour matching, and clustering were essential for detecting small red fruits. While fixed thresholding introduced day-to-day count variability, DBSCAN filtering and low‐degree polynomial smoothing recovered the overall fruiting trend and, when paired with wildlife visitation data, revealed clear canopy-then-ground foraging patterns. This hybrid strategy—combining off-the-shelf foundation models with system-aware, pixel-based metrics—delivers robust phenological insights without the need for large, hand-labeled datasets.

However, model variability can introduce noise into our ecological estimates through false positives, missed detections, and unreliable depth maps under dense foliage or low-light conditions. Figure~\ref{fig:oheloflowers} illustrates that the OWL detection + filtering method would miss many bird visits captured in very few frames. Statistical analysis confirms that the species composition of detected versus missed visits differs significantly ($\chi^2$, $p<0.005$), with missed observations more often involving amakihi than apapane. Such variability can markedly skew ecological conclusions, and model selection should always account for its potential impact on downstream analyses.


These findings also underscore several challenges in aligning camera-trap metrics with NEON's ground-based phenology records. For example, NEON reports percent‐open flowers, whereas our computer‐vision pipeline counts individual blossoms—two measures that do not map directly onto one another. NEON's use of coarse bins (e.g., logarithmic scales for berry counts) can further mask subtle trends that our high‐frequency imagery reveals. Moreover, each camera captures only a portion of a plant, while NEON's field surveys assess the entire individual. This partial view likely explains why our greenness curves often peak or plateau slightly before the manual records. To bridge these gaps, future work might deploy multiple cameras around each plant or adopt randomized camera orientations to better sample whole‐plant phenology. The early bird‐visit peak in Figure \ref{fig:oheloflowers} could possibly be because we placed cameras on parts of the plant that flowered earlier than the rest. While this "good view" placement enhances detection, it may bias timing estimates toward faster‐developing parts.

Despite these limitations, combining phenology records with camera-trap visitation data lets us develop hypotheses about species-specific interactions and the drivers of phenological change. We observed a clear point of inflection in pūkiawe berry counts around February 20 (Figure \ref{fig:berries}), which was followed by a rise in bird visits. The decrease in berry counts suggests that berry production for this species had slowed as dispersal intensified. Likewise, trends in 'ōhelo avian visitor species (Figure \ref{fig:oheloflowers}) reveal differences in foraging timing that mirror flower development and longevity, generating hypotheses of behavioral ecology and decision making among nectarivorous Hawaiian honey-creepers. Together, these findings show that fine-grained phenology metrics complement animal detections to uncover interaction dynamics that visitation data alone would miss. 

By integrating foundational AI models with targeted classical computer-vision methods on low-cost, standardized equipment, we establish a scalable framework for in situ monitoring of individual phenology and wildlife interactions. These results highlight the potential to produce standardized, high-resolution ecological datasets from readily available models and hardware, and develop new ecological observations and hypotheses. Further work will extend these methods to more species and phenological stages, with exciting implications for fine grained studies of phenology and ecological interactions. By lowering financial and technical barriers, these approaches also would allow for the broader application of automated monitoring protocols across diverse ecosystems.

\section{Acknowledgements}

This work was supported by the National Science Foundation OAC 2118240 Imageomics Institute and OISE 2330423 AI and Biodiversity Change Global Center awards and the Natural Sciences and Engineering Research Council of Canada 585136 award. This material draws on research supported in part by the Social Sciences and Humanities Research Council. Additionally, the material is based in part upon work supported by the National Ecological Observatory Network (NEON), a program sponsored by the U.S. National Science Foundation (NSF) and operated under cooperative agreement by Battelle.  Additionally, support for LM is provided by NSF award 2318597, and support for AP is provided by NSF award 2112606.

\bibliographystyle{plainnat}
\bibliography{references}

@article{tang2016emerging,
  title={Emerging opportunities and challenges in phenology: a review},
  author={Tang, Jianwu and K{\"o}rner, Christian and Muraoka, Hiroyuki and Piao, Shilong and Shen, Miaogen and Thackeray, Stephen J and Yang, Xi},
  journal={Ecosphere},
  volume={7},
  number={8},
  pages={e01436},
  year={2016},
  publisher={Wiley Online Library}
}

@article{piao2019plant,
  title={Plant phenology and global climate change: Current progresses and challenges},
  author={Piao, Shilong and Liu, Qiang and Chen, Anping and Janssens, Ivan A and Fu, Yongshuo and Dai, Junhu and Liu, Lingli and Lian, XU and Shen, Miaogen and Zhu, Xiaolin},
  journal={Global change biology},
  volume={25},
  number={6},
  pages={1922--1940},
  year={2019},
  publisher={Wiley Online Library}
}

@article{elmendorf2016plant,
  title={The plant phenology monitoring design for the National Ecological Observatory Network},
  author={Elmendorf, Sarah C and Jones, Katherine D and Cook, Benjamin I and Diez, Jeffrey M and Enquist, Carolyn AF and Hufft, Rebecca A and Jones, Matthew O and Mazer, Susan J and Miller-Rushing, Abraham J and Moore, David JP and others},
  journal={Ecosphere},
  volume={7},
  number={4},
  pages={e01303},
  year={2016},
  publisher={Wiley Online Library}
}

@article{beery2019efficient,
  title={Efficient pipeline for camera trap image review},
  author={Beery, Sara and Morris, Dan and Yang, Siyu},
  journal={arXiv preprint arXiv:1907.06772},
  year={2019}
}

@misc{dnr_2013, url={https://dlnr.hawaii.gov/ecosystems/files/2013/07/Puu-Makaala-Fact-Sheet.pdf}, journal={Puu-Makaala-Fact-Sheet}, author={Hawaii Department of Natural Resources}, year={2013}, month={Jul}}

@phdthesis{becker2024evolution,
  title={Evolution of Hawaiian Blueberries and Their Relatives},
  author={Becker, Anna},
  year={2024},
  school={University of Florida}
}

@article{mcinnes2018umap,
  title={Umap: Uniform manifold approximation and projection for dimension reduction},
  author={McInnes, Leland and Healy, John and Melville, James},
  journal={arXiv preprint arXiv:1802.03426},
  year={2018}
}

@inproceedings{ester1996density,
  title={A density-based algorithm for discovering clusters in large spatial databases with noise},
  author={Ester, Martin and Kriegel, Hans-Peter and Sander, J{\"o}rg and Xu, Xiaowei and others},
  booktitle={kdd},
  series={96},
  number={34},
  pages={226--231},
  year={1996}
}

@article{inouye2022climate,
  title={Climate change and phenology},
  author={Inouye, David W},
  journal={Wiley Interdisciplinary Reviews: Climate Change},
  volume={13},
  number={3},
  pages={e764},
  year={2022},
  publisher={Wiley Online Library}
}

@article{monteza2022arboreal,
  title={Arboreal camera trapping sheds light on seed dispersal of the world’s only epiphytic gymnosperm: Zamia pseudoparasitica},
  author={Monteza-Moreno, Claudio M and Rodriguez-Castro, Lilisbeth and Castillo-Caballero, Pedro L and Toribio, Edgar and Saltonstall, Kristin},
  journal={Ecology and Evolution},
  volume={12},
  number={3},
  pages={e8769},
  year={2022},
  publisher={Wiley Online Library}
}

@article{krauss2018effectiveness,
  title={Effectiveness of camera traps for quantifying daytime and nighttime visitation by vertebrate pollinators},
  author={Krauss, Siegfried L and Roberts, David G and Phillips, Ryan D and Edwards, Caroline},
  journal={Ecology and Evolution},
  volume={8},
  number={18},
  pages={9304--9314},
  year={2018},
  publisher={Wiley Online Library}
}

@article{sun2021simultaneous,
  title={Simultaneous monitoring of vegetation dynamics and wildlife activity with camera traps to assess habitat change},
  author={Sun, Catherine and Beirne, Christopher and Burgar, Joanna M and Howey, Thomas and Fisher, Jason T and Burton, A Cole},
  journal={Remote Sensing in Ecology and Conservation},
  volume={7},
  number={4},
  pages={666--684},
  year={2021},
  publisher={Wiley Online Library}
}

@article{de2022camera,
  title={Camera traps enable the estimation of herbaceous aboveground net primary production (ANPP) in an African savanna at high temporal resolution},
  author={de Jonge, Inger K and Veldhuis, Michiel P and Vrieling, Anton and Olff, Han},
  journal={Remote Sensing in Ecology and Conservation},
  volume={8},
  number={5},
  pages={583--600},
  year={2022},
  publisher={Wiley Online Library}
}

@article{hofmeester2020using,
  title={Using by-catch data from wildlife surveys to quantify climatic parameters and timing of phenology for plants and animals using camera traps},
  author={Hofmeester, Tim R and Young, Sherry and Juthberg, Sonya and Singh, Navinder J and Widemo, Fredrik and Andr{\'e}n, Henrik and Linnell, John DC and Cromsigt, Joris PGM},
  journal={Remote Sensing in Ecology and Conservation},
  volume={6},
  number={2},
  pages={129--140},
  year={2020},
  publisher={Wiley Online Library}
}

@article{villalva2024frugivory,
  title={FRUGIVORY CAMTRAP: A dataset of plant-animal interactions recorded with camera traps},
  author={Villalva, Pablo and Arroyo-Correa, Blanca and Calvo, Gemma and Homet, Pablo and Isla, Jorge and Mendoza, Irene and Moracho, Eva and Quintero, Elena and Rodr{\'\i}guez-S{\'a}nchez, Francisco and Jordano, Pedro},
  journal={Ecology},
  volume={105},
  number={11},
  pages={e4424},
  year={2024},
  publisher={John Wiley \& Sons, Inc. Hoboken, USA}
}

@article{ovaskainen2013community,
  title={Community-level phenological response to climate change},
  author={Ovaskainen, Otso and Skorokhodova, Svetlana and Yakovleva, Marina and Sukhov, Alexander and Kutenkov, Anatoliy and Kutenkova, Nadezhda and Shcherbakov, Anatoliy and Meyke, Evegeniy and Delgado, Maria del Mar},
  journal={Proceedings of the National Academy of Sciences},
  volume={110},
  number={33},
  pages={13434--13439},
  year={2013},
  publisher={National Academy of Sciences}
}

@article{morisette2009tracking,
  title={Tracking the rhythm of the seasons in the face of global change: phenological research in the 21st century},
  author={Morisette, Jeffrey T and Richardson, Andrew D and Knapp, Alan K and Fisher, Jeremy I and Graham, Eric A and Abatzoglou, John and Wilson, Bruce E and Breshears, David D and Henebry, Geoffrey M and Hanes, Jonathan M and others},
  journal={Frontiers in Ecology and the Environment},
  volume={7},
  number={5},
  pages={253--260},
  year={2009},
  publisher={Wiley Online Library}
}

@article{chuine2010does,
  title={Why does phenology drive species distribution?},
  author={Chuine, Isabelle},
  journal={Philosophical Transactions of the Royal Society B: Biological Sciences},
  volume={365},
  number={1555},
  pages={3149--3160},
  year={2010},
  publisher={The Royal Society}
}

@article{keller2008continental,
  title={A continental strategy for the National Ecological Observatory Network},
  author={Keller, Michael and Schimel, David S and Hargrove, W William and Hoffman, Forrest M},
  journal={The Ecological Society of America: 282-284},
  year={2008}
}

@article{richardson2018tracking,
  title={Tracking vegetation phenology across diverse North American biomes using PhenoCam imagery},
  author={Richardson, Andrew D and Hufkens, Koen and Milliman, Tom and Aubrecht, Donald M and Chen, Min and Gray, Josh M and Johnston, Miriam R and Keenan, Trevor F and Klosterman, Stephen T and Kosmala, Margaret and others},
  journal={Scientific data},
  volume={5},
  number={1},
  pages={1--24},
  year={2018},
  publisher={Nature Publishing Group}
}

@article{deitke2024molmo,
  title={Molmo and pixmo: Open weights and open data for state-of-the-art multimodal models},
  author={Deitke, Matt and Clark, Christopher and Lee, Sangho and Tripathi, Rohun and Yang, Yue and Park, Jae Sung and Salehi, Mohammadreza and Muennighoff, Niklas and Lo, Kyle and Soldaini, Luca and others},
  journal={arXiv preprint arXiv:2409.17146},
  year={2024}
}

@article{crimmins2008monitoring,
  title={Monitoring plant phenology using digital repeat photography},
  author={Crimmins, Michael A and Crimmins, Theresa M},
  journal={Environmental management},
  volume={41},
  pages={949--958},
  year={2008},
  publisher={Springer}
}

@article{pau2020climatic,
  title={Climatic sensitivity of species’ vegetative and reproductive phenology in a Hawaiian montane wet forest},
  author={Pau, Stephanie and Cordell, Susan and Ostertag, Rebecca and Inman, Faith and Sack, Lawren},
  journal={Biotropica},
  volume={52},
  number={5},
  pages={825--835},
  year={2020},
  publisher={Wiley Online Library}
}

@inproceedings{
dosovitskiy2021an,
title={An Image is Worth 16x16 Words: Transformers for Image Recognition at Scale},
author={Alexey Dosovitskiy and Lucas Beyer and Alexander Kolesnikov and Dirk Weissenborn and Xiaohua Zhai and Thomas Unterthiner and Mostafa Dehghani and Matthias Minderer and Georg Heigold and Sylvain Gelly and Jakob Uszkoreit and Neil Houlsby},
booktitle={International Conference on Learning Representations},
year={2021},
url={https://openreview.net/forum?id=YicbFdNTTy}
}

@inproceedings{NIPS2017_3f5ee243,
 author = {Vaswani, Ashish and Shazeer, Noam and Parmar, Niki and Uszkoreit, Jakob and Jones, Llion and Gomez, Aidan N and Kaiser, \L ukasz and Polosukhin, Illia},
 booktitle = {Advances in Neural Information Processing Systems},
 editor = {I. Guyon and U. Von Luxburg and S. Bengio and H. Wallach and R. Fergus and S. Vishwanathan and R. Garnett},
 pages = {},
 publisher = {Curran Associates, Inc.},
 title = {Attention is All you Need},
 url = {https://proceedings.neurips.cc/paper_files/paper/2017/file/3f5ee243547dee91fbd053c1c4a845aa-Paper.pdf},
 volume = {30},
 year = {2017}
}

@InProceedings{pmlr-v139-radford21a,
  title = 	 {Learning Transferable Visual Models From Natural Language Supervision},
  author =       {Radford, Alec and Kim, Jong Wook and Hallacy, Chris and Ramesh, Aditya and Goh, Gabriel and Agarwal, Sandhini and Sastry, Girish and Askell, Amanda and Mishkin, Pamela and Clark, Jack and Krueger, Gretchen and Sutskever, Ilya},
  booktitle = 	 {Proceedings of the 38th International Conference on Machine Learning},
  pages = 	 {8748--8763},
  year = 	 {2021},
  editor = 	 {Meila, Marina and Zhang, Tong},
  volume = 	 {139},
  series = 	 {Proceedings of Machine Learning Research},
  month = 	 {18--24 Jul},
  publisher =    {PMLR},
}

@inproceedings{DBLP:conf/icml/JiaYXCPPLSLD21,
  author       = {Chao Jia and
                  Yinfei Yang and
                  Ye Xia and
                  Yi{-}Ting Chen and
                  Zarana Parekh and
                  Hieu Pham and
                  Quoc V. Le and
                  Yun{-}Hsuan Sung and
                  Zhen Li and
                  Tom Duerig},
  editor       = {Marina Meila and
                  Tong Zhang},
  title        = {Scaling Up Visual and Vision-Language Representation Learning With
                  Noisy Text Supervision},
  booktitle    = {Proceedings of the 38th International Conference on Machine Learning,
                  {ICML} 2021, 18-24 July 2021, Virtual Event},
  series       = {Proceedings of Machine Learning Research},
  volume       = {139},
  pages        = {4904--4916},
  publisher    = {{PMLR}},
  year         = {2021},
  url          = {http://proceedings.mlr.press/v139/jia21b.html},
  bibsource    = {dblp computer science bibliography, https://dblp.org}
}

@inproceedings{10.5555/3600270.3601993,
author = {Alayrac, Jean-Baptiste and Donahue, Jeff and Luc, Pauline and Miech, Antoine and Barr, Iain and Hasson, Yana and Lenc, Karel and Mensch, Arthur and Millicah, Katie and Reynolds, Malcolm and Ring, Roman and Rutherford, Eliza and Cabi, Serkan and Han, Tengda and Gong, Zhitao and Samangooei, Sina and Monteiro, Marianne and Menick, Jacob and Borgeaud, Sebastian and Brock, Andrew and Nematzadeh, Aida and Sharifzadeh, Sahand and Binkowski, Mikolaj and Barreira, Ricardo and Vinyals, Oriol and Zisserman, Andrew and Simonyan, Karen},
title = {Flamingo: a visual language model for few-shot learning},
year = {2022},
isbn = {9781713871088},
publisher = {Curran Associates Inc.},
address = {Red Hook, NY, USA},
booktitle = {Proceedings of the 36th International Conference on Neural Information Processing Systems},
articleno = {1723},
numpages = {21},
location = {New Orleans, LA, USA},
series = {NIPS '22}
}

@ARTICLE{2023arXiv230308774O,
       author = {{OpenAI} and {Achiam}, Josh and {Adler}, Steven and {Agarwal}, Sandhini and {Ahmad}, Lama and {Akkaya}, Ilge and {Leoni Aleman}, Florencia and {Almeida}, Diogo and {Altenschmidt}, Janko and {Altman}, Sam and {Anadkat}, Shyamal and {Avila}, Red and {Babuschkin}, Igor and {Balaji}, Suchir and {Balcom}, Valerie and {Baltescu}, Paul and {Bao}, Haiming and {Bavarian}, Mohammad and {Belgum}, Jeff and {Bello}, Irwan and {Berdine}, Jake and {Bernadett-Shapiro}, Gabriel and {Berner}, Christopher and {Bogdonoff}, Lenny and {Boiko}, Oleg and {Boyd}, Madelaine and {Brakman}, Anna-Luisa and {Brockman}, Greg and {Brooks}, Tim and {Brundage}, Miles and {Button}, Kevin and {Cai}, Trevor and {Campbell}, Rosie and {Cann}, Andrew and {Carey}, Brittany and {Carlson}, Chelsea and {Carmichael}, Rory and {Chan}, Brooke and {Chang}, Che and {Chantzis}, Fotis and {Chen}, Derek and {Chen}, Sully and {Chen}, Ruby and {Chen}, Jason and {Chen}, Mark and {Chess}, Ben and {Cho}, Chester and {Chu}, Casey and {Chung}, Hyung Won and {Cummings}, Dave and {Currier}, Jeremiah and {Dai}, Yunxing and {Decareaux}, Cory and {Degry}, Thomas and {Deutsch}, Noah and {Deville}, Damien and {Dhar}, Arka and {Dohan}, David and {Dowling}, Steve and {Dunning}, Sheila and {Ecoffet}, Adrien and {Eleti}, Atty and {Eloundou}, Tyna and {Farhi}, David and {Fedus}, Liam and {Felix}, Niko and {Posada Fishman}, Sim{\'o}n and {Forte}, Juston and {Fulford}, Isabella and {Gao}, Leo and {Georges}, Elie and {Gibson}, Christian and {Goel}, Vik and {Gogineni}, Tarun and {Goh}, Gabriel and {Gontijo-Lopes}, Rapha and {Gordon}, Jonathan and {Grafstein}, Morgan and {Gray}, Scott and {Greene}, Ryan and {Gross}, Joshua and {Gu}, Shixiang Shane and {Guo}, Yufei and {Hallacy}, Chris and {Han}, Jesse and {Harris}, Jeff and {He}, Yuchen and {Heaton}, Mike and {Heidecke}, Johannes and {Hesse}, Chris and {Hickey}, Alan and {Hickey}, Wade and {Hoeschele}, Peter and {Houghton}, Brandon and {Hsu}, Kenny and {Hu}, Shengli and {Hu}, Xin and {Huizinga}, Joost and {Jain}, Shantanu and {Jain}, Shawn and {Jang}, Joanne and {Jiang}, Angela and {Jiang}, Roger and {Jin}, Haozhun and {Jin}, Denny and {Jomoto}, Shino and {Jonn}, Billie and {Jun}, Heewoo and {Kaftan}, Tomer and {Kaiser}, {\L}ukasz and {Kamali}, Ali and {Kanitscheider}, Ingmar and {Shirish Keskar}, Nitish and {Khan}, Tabarak and {Kilpatrick}, Logan and {Kim}, Jong Wook and {Kim}, Christina and {Kim}, Yongjik and {Hendrik Kirchner}, Jan and {Kiros}, Jamie and {Knight}, Matt and {Kokotajlo}, Daniel and {Kondraciuk}, {\L}ukasz and {Kondrich}, Andrew and {Konstantinidis}, Aris and {Kosic}, Kyle and {Krueger}, Gretchen and {Kuo}, Vishal and {Lampe}, Michael and {Lan}, Ikai and {Lee}, Teddy and {Leike}, Jan and {Leung}, Jade and {Levy}, Daniel and {Li}, Chak Ming and {Lim}, Rachel and {Lin}, Molly and {Lin}, Stephanie and {Litwin}, Mateusz and {Lopez}, Theresa and {Lowe}, Ryan and {Lue}, Patricia and {Makanju}, Anna and {Malfacini}, Kim and {Manning}, Sam and {Markov}, Todor and {Markovski}, Yaniv and {Martin}, Bianca and {Mayer}, Katie and {Mayne}, Andrew and {McGrew}, Bob and {McKinney}, Scott Mayer and {McLeavey}, Christine and {McMillan}, Paul and {McNeil}, Jake and {Medina}, David and {Mehta}, Aalok and {Menick}, Jacob and {Metz}, Luke and {Mishchenko}, Andrey and {Mishkin}, Pamela and {Monaco}, Vinnie and {Morikawa}, Evan and {Mossing}, Daniel and {Mu}, Tong and {Murati}, Mira and {Murk}, Oleg and {M{\'e}ly}, David and {Nair}, Ashvin and {Nakano}, Reiichiro and {Nayak}, Rajeev and {Neelakantan}, Arvind and {Ngo}, Richard and {Noh}, Hyeonwoo and {Ouyang}, Long and {O'Keefe}, Cullen and {Pachocki}, Jakub and {Paino}, Alex and {Palermo}, Joe and {Pantuliano}, Ashley and {Parascandolo}, Giambattista and {Parish}, Joel and {Parparita}, Emy and {Passos}, Alex and {Pavlov}, Mikhail and {Peng}, Andrew and {Perelman}, Adam and {de Avila Belbute Peres}, Filipe and {Petrov}, Michael and {Ponde de Oliveira Pinto}, Henrique and {Michael} and {Pokorny} and {Pokrass}, Michelle and {Pong}, Vitchyr H. and {Powell}, Tolly and {Power}, Alethea and {Power}, Boris and {Proehl}, Elizabeth and {Puri}, Raul and {Radford}, Alec},
        title = "{GPT-4 Technical Report}",
      journal = {arXiv e-prints},
         year = 2023,
        month = mar,
          eid = {arXiv:2303.08774},
        pages = {arXiv:2303.08774},
          doi = {10.48550/arXiv.2303.08774},
archivePrefix = {arXiv},
       eprint = {2303.08774},
 primaryClass = {cs.CL},
       adsurl = {https://ui.adsabs.harvard.edu/abs/2023arXiv230308774O}
}

@INPROCEEDINGS{10377550,
  author={Zhai, Xiaohua and Mustafa, Basil and Kolesnikov, Alexander and Beyer, Lucas},
  booktitle={2023 IEEE/CVF International Conference on Computer Vision (ICCV)}, 
  title={Sigmoid Loss for Language Image Pre-Training}, 
  year={2023},
  volume={},
  number={},
  pages={11941-11952},
  doi={10.1109/ICCV51070.2023.01100}}

@inproceedings{10.1007/978-3-031-72970-6_3,
author = {Liu, Shilong and Zeng, Zhaoyang and Ren, Tianhe and Li, Feng and Zhang, Hao and Yang, Jie and Jiang, Qing and Li, Chunyuan and Yang, Jianwei and Su, Hang and Zhu, Jun and Zhang, Lei},
title = {Grounding DINO: Marrying DINO with Grounded Pre-training for Open-Set Object Detection},
year = {2024},
isbn = {978-3-031-72969-0},
publisher = {Springer-Verlag},
address = {Berlin, Heidelberg},
url = {https://doi.org/10.1007/978-3-031-72970-6_3},
doi = {10.1007/978-3-031-72970-6_3},
booktitle = {Computer Vision – ECCV 2024: 18th European Conference, Milan, Italy, September 29–October 4, 2024, Proceedings, Part XLVII},
pages = {38–55},
numpages = {18},
location = {Milan, Italy}
}

@INPROCEEDINGS{10657519,
  author={Xiao, Bin and Wu, Haiping and Xu, Weijian and Dai, Xiyang and Hu, Houdong and Lu, Yumao and Zeng, Michael and Liu, Ce and Yuan, Lu},
  booktitle={2024 IEEE/CVF Conference on Computer Vision and Pattern Recognition (CVPR)}, 
  title={Florence-2: Advancing a Unified Representation for a Variety of Vision Tasks}, 
  year={2024},
  volume={},
  number={},
  pages={4818-4829},
  doi={10.1109/CVPR52733.2024.00461}}

@inproceedings{10.5555/3666122.3669313,
author = {Minderer, Matthias and Gritsenko, Alexey and Houlsby, Neil},
title = {Scaling open-vocabulary object detection},
year = {2023},
publisher = {Curran Associates Inc.},
address = {Red Hook, NY, USA},
booktitle = {Proceedings of the 37th International Conference on Neural Information Processing Systems},
articleno = {3191},
numpages = {25},
location = {New Orleans, LA, USA},
series = {NIPS '23}
}

@inproceedings{
ravi2025sam,
title={{SAM} 2: Segment Anything in Images and Videos},
author={Nikhila Ravi and Valentin Gabeur and Yuan-Ting Hu and Ronghang Hu and Chaitanya Ryali and Tengyu Ma and Haitham Khedr and Roman R{\"a}dle and Chloe Rolland and Laura Gustafson and Eric Mintun and Junting Pan and Kalyan Vasudev Alwala and Nicolas Carion and Chao-Yuan Wu and Ross Girshick and Piotr Dollar and Christoph Feichtenhofer},
booktitle={The Thirteenth International Conference on Learning Representations},
year={2025},
url={https://openreview.net/forum?id=Ha6RTeWMd0}
}

@article{
wang2022git,
title={{GIT}: A Generative Image-to-text Transformer for Vision and Language},
author={Jianfeng Wang and Zhengyuan Yang and Xiaowei Hu and Linjie Li and Kevin Lin and Zhe Gan and Zicheng Liu and Ce Liu and Lijuan Wang},
journal={Transactions on Machine Learning Research},
issn={2835-8856},
year={2022},
url={https://openreview.net/forum?id=b4tMhpN0JC},
note={}
}

@inproceedings{10.5555/3618408.3619222,
author = {Li, Junnan and Li, Dongxu and Savarese, Silvio and Hoi, Steven},
title = {BLIP-2: bootstrapping language-image pre-training with frozen image encoders and large language models},
year = {2023},
publisher = {JMLR.org},
booktitle = {Proceedings of the 40th International Conference on Machine Learning},
articleno = {814},
numpages = {13},
location = {Honolulu, Hawaii, USA},
series = {ICML'23}
}

@InProceedings{Suma_2024_ECCV,
    author    = {Suma, Pavel and Kordopatis-Zilos, Giorgos and Iscen, Ahmet and Tolias, Giorgos},
    title     = {AMES: Asymmetric and Memory-Efficient Similarity Estimation for Instance-level Retrieval},
    booktitle = {European Conference on Computer Vision (ECCV)},
    year      = {2024}
}

@INPROCEEDINGS{10378372,
  author={Shao, Shihao and Chen, Kaifeng and Karpur, Arjun and Cui, Qinghua and Araujo, André and Cao, Bingyi},
  booktitle={2023 IEEE/CVF International Conference on Computer Vision (ICCV)}, 
  title={Global Features are All You Need for Image Retrieval and Reranking}, 
  year={2023},
  volume={},
  number={},
  pages={11002-11012},
  doi={10.1109/ICCV51070.2023.01013}}

@article{li2019visualbert,
  title={Visualbert: A simple and performant baseline for vision and language},
  author={Li, Liunian Harold and Yatskar, Mark and Yin, Da and Hsieh, Cho-Jui and Chang, Kai-Wei},
  journal={arXiv preprint arXiv:1908.03557},
  year={2019}
}

@inproceedings{wang2023image,
  title={Image as a foreign language: Beit pretraining for vision and vision-language tasks},
  author={Wang, Wenhui and Bao, Hangbo and Dong, Li and Bjorck, Johan and Peng, Zhiliang and Liu, Qiang and Aggarwal, Kriti and Mohammed, Owais Khan and Singhal, Saksham and Som, Subhojit and others},
  booktitle={Proceedings of the IEEE/CVF Conference on Computer Vision and Pattern Recognition},
  pages={19175--19186},
  year={2023}
}

@article{team2023gemini,
  title={Gemini: a family of highly capable multimodal models},
  author={Team, Gemini and Anil, Rohan and Borgeaud, Sebastian and Alayrac, Jean-Baptiste and Yu, Jiahui and Soricut, Radu and Schalkwyk, Johan and Dai, Andrew M and Hauth, Anja and Millican, Katie and others},
  journal={arXiv preprint arXiv:2312.11805},
  year={2023}
}

@article{bai2025qwen2,
  title={Qwen2. 5-vl technical report},
  author={Bai, Shuai and Chen, Keqin and Liu, Xuejing and Wang, Jialin and Ge, Wenbin and Song, Sibo and Dang, Kai and Wang, Peng and Wang, Shijie and Tang, Jun and others},
  journal={arXiv preprint arXiv:2502.13923},
  year={2025}
}

@article{https://doi.org/10.1111/j.1469-8137.2011.03803.x,
author = {Polgar, Caroline A. and Primack, Richard B.},
title = {Leaf-out phenology of temperate woody plants: from trees to ecosystems},
journal = {New Phytologist},
volume = {191},
number = {4},
pages = {926-941},
doi = {https://doi.org/10.1111/j.1469-8137.2011.03803.x},
url = {https://nph.onlinelibrary.wiley.com/doi/abs/10.1111/j.1469-8137.2011.03803.x},
eprint = {https://nph.onlinelibrary.wiley.com/doi/pdf/10.1111/j.1469-8137.2011.03803.x},
year = {2011}
}

@article{https://doi.org/10.1111/nph.14255,
author = {Gilliam, Frank S.},
title = {Forest ecosystems of temperate climatic regions: from ancient use to climate change},
journal = {New Phytologist},
volume = {212},
number = {4},
pages = {871-887},
doi = {https://doi.org/10.1111/nph.14255},
url = {https://nph.onlinelibrary.wiley.com/doi/abs/10.1111/nph.14255},
eprint = {https://nph.onlinelibrary.wiley.com/doi/pdf/10.1111/nph.14255},
year = {2016}
}

@article{10.1093/biosci/biz063,
    author = {Culumber, Zachary W and Anaya-Rojas, Jaime M and Booker, William W and Hooks, Alexandra P and Lange, Elizabeth C and Pluer, Benjamin and Ramírez-Bullón, Natali and Travis, Joseph},
    title = {Widespread Biases in Ecological and Evolutionary Studies},
    journal = {BioScience},
    volume = {69},
    number = {8},
    pages = {631-640},
    year = {2019},
    month = {07},
    issn = {0006-3568},
    doi = {10.1093/biosci/biz063},
    url = {https://doi.org/10.1093/biosci/biz063},
    eprint = {https://academic.oup.com/bioscience/article-pdf/69/8/631/29099526/biz063.pdf},
}

@article{https://doi.org/10.1890/ES12-00299.1,
author = {Trimble, Morgan J. and van Aarde, Rudi J.},
title = {Geographical and taxonomic biases in research on biodiversity in human-modified landscapes},
journal = {Ecosphere},
volume = {3},
number = {12},
pages = {art119},
doi = {https://doi.org/10.1890/ES12-00299.1},
url = {https://esajournals.onlinelibrary.wiley.com/doi/abs/10.1890/ES12-00299.1},
eprint = {https://esajournals.onlinelibrary.wiley.com/doi/pdf/10.1890/ES12-00299.1},
year = {2012}
}

@article{https://doi.org/10.1002/ecy.3846,
author = {Stemkovski, Michael and Bell, James R. and Ellwood, Elizabeth R. and Inouye, Brian D. and Kobori, Hiromi and Lee, Sang Don and Lloyd-Evans, Trevor and Primack, Richard B. and Templ, Barbara and Pearse, William D.},
title = {Disorder or a new order: How climate change affects phenological variability},
journal = {Ecology},
volume = {104},
number = {1},
pages = {e3846},
doi = {https://doi.org/10.1002/ecy.3846},
url = {https://esajournals.onlinelibrary.wiley.com/doi/abs/10.1002/ecy.3846},
eprint = {https://esajournals.onlinelibrary.wiley.com/doi/pdf/10.1002/ecy.3846},
year = {2023}
}

@article{doi:10.1139/X08-104,
author = {Campbell, John L. and Rustad, Lindsey E. and Boyer, Elizabeth W. and Christopher, Sheila F. and Driscoll, Charles T. and Fernandez, Ivan J. and Groffman, Peter M. and Houle, Daniel and Kiekbusch, Jana and Magill, Alison H. and Mitchell, Myron J. and Ollinger, Scott V.},
title = {Consequences of climate change for biogeochemical cycling in forests of northeastern North AmericaThis article is one of a selection of papers from NE Forests 2100: A Synthesis of Climate Change Impacts on Forests of the Northeastern US and Eastern Canada.},
journal = {Canadian Journal of Forest Research},
volume = {39},
number = {2},
pages = {264-284},
year = {2009},
doi = {10.1139/X08-104},
URL = {https://doi.org/10.1139/X08-104},
eprint = {https://doi.org/10.1139/X08-104}
}

@article{PARK2021709,
title = {Scale gaps in landscape phenology: challenges and opportunities},
journal = {Trends in Ecology \& Evolution},
volume = {36},
number = {8},
pages = {709-721},
year = {2021},
issn = {0169-5347},
doi = {https://doi.org/10.1016/j.tree.2021.04.008},
url = {https://www.sciencedirect.com/science/article/pii/S0169534721001087},
author = {Daniel S. Park and Erica A. Newman and Ian K. Breckheimer}
}

@article{tanwar2021camera,
  title={Camera trap placement for evaluating species richness, abundance, and activity},
  author={Tanwar, Kamakshi S and Sadhu, Ayan and Jhala, Yadvendradev V},
  journal={Scientific Reports},
  volume={11},
  number={1},
  pages={23050},
  year={2021},
  publisher={Nature Publishing Group UK London}
}

@ARTICLE{10.3389/fevo.2021.617996,
AUTHOR={Delisle, Zackary J.  and Flaherty, Elizabeth A.  and Nobbe, Mackenzie R.  and Wzientek, Cole M.  and Swihart, Robert K. },    
TITLE={Next-Generation Camera Trapping: Systematic Review of Historic Trends Suggests Keys to Expanded Research Applications in Ecology and Conservation},     
JOURNAL={Frontiers in Ecology and Evolution},
VOLUME={Volume 9 - 2021},
YEAR={2021},
URL={https://www.frontiersin.org/journals/ecology-and-evolution/articles/10.3389/fevo.2021.617996},
DOI={10.3389/fevo.2021.617996},
ISSN={2296-701X}}

@article{JAVAID2024792,
title = {Computer vision to enhance healthcare domain: An overview of features, implementation, and opportunities},
journal = {Intelligent Pharmacy},
volume = {2},
number = {6},
pages = {792-803},
year = {2024},
issn = {2949-866X},
doi = {https://doi.org/10.1016/j.ipha.2024.05.007},
url = {https://www.sciencedirect.com/science/article/pii/S2949866X24000662},
author = {Mohd Javaid and Abid Haleem and Ravi Pratap Singh and Mumtaz Ahmed}
}

@article{HARIPAVAN2025100371,
title = {Integration of geospatial techniques and machine learning in land parcel prediction},
journal = {Geosystems and Geoenvironment},
volume = {4},
number = {2},
pages = {100371},
year = {2025},
issn = {2772-8838},
doi = {https://doi.org/10.1016/j.geogeo.2025.100371},
url = {https://www.sciencedirect.com/science/article/pii/S2772883825000214},
author = {Nekkanti Haripavan and Subhashish Dey and Chimakurthi Harika Mani Chandana}
}

@Article{rs17020179,
AUTHOR = {Huo, Chunlei and Chen, Keming and Zhang, Shuaihao and Wang, Zeyu and Yan, Heyu and Shen, Jing and Hong, Yuyang and Qi, Geqi and Fang, Hongmei and Wang, Zihan},
TITLE = {When Remote Sensing Meets Foundation Model: A Survey and Beyond},
JOURNAL = {Remote Sensing},
VOLUME = {17},
YEAR = {2025},
NUMBER = {2},
ARTICLE-NUMBER = {179},
URL = {https://www.mdpi.com/2072-4292/17/2/179},
ISSN = {2072-4292},
DOI = {10.3390/rs17020179}
}

@INPROCEEDINGS{9144058,
  author={TOMBE, Ronald},
  booktitle={2020 IST-Africa Conference (IST-Africa)}, 
  title={Computer Vision for Smart Farming and Sustainable Agriculture}, 
  year={2020},
  volume={},
  number={},
  pages={1-8},
  doi={}
}

@article{10.1145/3626186,
author = {Luo, Jiayun and Li, Boyang and Leung, Cyril},
title = {A Survey of Computer Vision Technologies in Urban and Controlled-environment Agriculture},
year = {2023},
issue_date = {May 2024},
publisher = {Association for Computing Machinery},
address = {New York, NY, USA},
volume = {56},
number = {5},
issn = {0360-0300},
url = {https://doi.org/10.1145/3626186},
doi = {10.1145/3626186},
journal = {ACM Comput. Surv.},
month = nov,
articleno = {118},
numpages = {39}
}

@article{DHANYA2022211,
title = {Deep learning based computer vision approaches for smart agricultural applications},
journal = {Artificial Intelligence in Agriculture},
volume = {6},
pages = {211-229},
year = {2022},
issn = {2589-7217},
doi = {https://doi.org/10.1016/j.aiia.2022.09.007},
url = {https://www.sciencedirect.com/science/article/pii/S2589721722000174},
author = {V.G. Dhanya and A. Subeesh and N.L. Kushwaha and Dinesh Kumar Vishwakarma and T. {Nagesh Kumar} and G. Ritika and A.N. Singh}
}

@Article{Chandra2021,
author={Chandra, Mayank Arya
and Bedi, S. S.},
title={Survey on SVM and their application in imageclassification},
journal={International Journal of Information Technology},
year={2021},
month={Oct},
day={01},
volume={13},
number={5},
pages={1-11},
issn={2511-2112},
doi={10.1007/s41870-017-0080-1},
url={https://doi.org/10.1007/s41870-017-0080-1}
}

@ARTICLE{6975001,
  author={Khurshid, Hasnat and Khan, Muhammad Faisal},
  journal={IEEE Journal of Selected Topics in Applied Earth Observations and Remote Sensing}, 
  title={Segmentation and Classification Using Logistic Regression in Remote Sensing Imagery}, 
  year={2015},
  volume={8},
  number={1},
  pages={224-232},
  doi={10.1109/JSTARS.2014.2362769}}

@Article{electronics12051199,
AUTHOR = {Yu, Ying and Wang, Chunping and Fu, Qiang and Kou, Renke and Huang, Fuyu and Yang, Boxiong and Yang, Tingting and Gao, Mingliang},
TITLE = {Techniques and Challenges of Image Segmentation: A Review},
JOURNAL = {Electronics},
VOLUME = {12},
YEAR = {2023},
NUMBER = {5},
ARTICLE-NUMBER = {1199},
URL = {https://www.mdpi.com/2079-9292/12/5/1199},
ISSN = {2079-9292},
DOI = {10.3390/electronics12051199}
}

@inproceedings{NIPS2012_c399862d,
 author = {Krizhevsky, Alex and Sutskever, Ilya and Hinton, Geoffrey E},
 booktitle = {Advances in Neural Information Processing Systems},
 editor = {F. Pereira and C.J. Burges and L. Bottou and K.Q. Weinberger},
 pages = {},
 publisher = {Curran Associates, Inc.},
 title = {ImageNet Classification with Deep Convolutional Neural Networks},
 url = {https://proceedings.neurips.cc/paper_files/paper/2012/file/c399862d3b9d6b76c8436e924a68c45b-Paper.pdf},
 volume = {25},
 year = {2012}
}

@InProceedings{He_2016_CVPR,
author = {He, Kaiming and Zhang, Xiangyu and Ren, Shaoqing and Sun, Jian},
title = {Deep Residual Learning for Image Recognition},
booktitle = {Proceedings of the IEEE Conference on Computer Vision and Pattern Recognition (CVPR)},
month = {June},
year = {2016}
}

@article{https://doi.org/10.1155/2018/7068349,
author = {Voulodimos, Athanasios and Doulamis, Nikolaos and Doulamis, Anastasios and Protopapadakis, Eftychios},
title = {Deep Learning for Computer Vision: A Brief Review},
journal = {Computational Intelligence and Neuroscience},
volume = {2018},
number = {1},
pages = {7068349},
doi = {https://doi.org/10.1155/2018/7068349},
url = {https://onlinelibrary.wiley.com/doi/abs/10.1155/2018/7068349},
eprint = {https://onlinelibrary.wiley.com/doi/pdf/10.1155/2018/7068349},
year = {2018}
}

@article{doi:10.1139/er-2018-0034,
author = {Liu, Zelin and Peng, Changhui and Work, Timothy and Candau, Jean-Noel and DesRochers, Annie and Kneeshaw, Daniel},
title = {Application of machine-learning methods in forest ecology: recent progress and future challenges},
journal = {Environmental Reviews},
volume = {26},
number = {4},
pages = {339-350},
year = {2018},
doi = {10.1139/er-2018-0034},URL = {https://doi.org/10.1139/er-2018-0034},
eprint = {https://doi.org/10.1139/er-2018-0034}
}

@Article{TalaeiKhoei2023,
author={Talaei Khoei, Tala
and Ould Slimane, Hadjar
and Kaabouch, Naima},
title={Deep learning: systematic review, models, challenges, and research directions},
journal={Neural Computing and Applications},
year={2023},
month={Nov},
day={01},
volume={35},
number={31},
pages={23103-23124},
issn={1433-3058},
doi={10.1007/s00521-023-08957-4},
url={https://doi.org/10.1007/s00521-023-08957-4}
}

@Article{Hosna2022,
author={Hosna, Asmaul
and Merry, Ethel
and Gyalmo, Jigmey
and Alom, Zulfikar
and Aung, Zeyar
and Azim, Mohammad Abdul},
title={Transfer learning: a friendly introduction},
journal={Journal of Big Data},
year={2022},
month={Oct},
day={22},
volume={9},
number={1},
pages={102},
issn={2196-1115},
doi={10.1186/s40537-022-00652-w},
url={https://doi.org/10.1186/s40537-022-00652-w}
}

@inproceedings{gupta2022deep,
  title={Deep learning (CNN) and transfer learning: a review},
  author={Gupta, Jaya and Pathak, Sunil and Kumar, Gireesh},
  booktitle={Journal of Physics: Conference Series},
  volume={2273},
  pages={012029},
  year={2022},
  organization={IOP Publishing}
}

@article{zhao2024comparison,
  title={A comparison review of transfer learning and self-supervised learning: Definitions, applications, advantages and limitations},
  author={Zhao, Zehui and Alzubaidi, Laith and Zhang, Jinglan and Duan, Ye and Gu, Yuantong},
  journal={Expert Systems with Applications},
  volume={242},
  pages={122807},
  year={2024},
  publisher={Elsevier}
}

@article{10.1145/3582688,
author = {Song, Yisheng and Wang, Ting and Cai, Puyu and Mondal, Subrota K. and Sahoo, Jyoti Prakash},
title = {A Comprehensive Survey of Few-shot Learning: Evolution, Applications, Challenges, and Opportunities},
year = {2023},
issue_date = {December 2023},
publisher = {Association for Computing Machinery},
address = {New York, NY, USA},
volume = {55},
number = {13s},
issn = {0360-0300},
url = {https://doi.org/10.1145/3582688},
doi = {10.1145/3582688},
journal = {ACM Comput. Surv.},
month = jul,
articleno = {271},
numpages = {40}
}

@article{quevedo2025pixel,
  title={Pixel-by-Pixel Analysis of Soil and Leaf Coverage in Purslane: A CIELAB Approach},
  author={Quevedo-Nolasco, Abel and Aguado-Rodr{\'\i}, Graciano-Javier and Lara-Viveros, Francisco-Marcelo and Landero-Valenzuela, Nadia and others},
  journal={Agricultural Sciences},
  volume={16},
  number={2},
  pages={227--239},
  year={2025},
  publisher={Scientific Research Publishing}
}

@ARTICLE{10136612,
  author={He, Longke and Cheng, Xiao and Jiwa, Aying and Li, Dan and Fang, Jing and Du, Zhencong},
  journal={IEEE Sensors Journal}, 
  title={Zanthoxylum bungeanum Fruit Detection by Adaptive Thresholds in HSV Space for an Automatic Picking System}, 
  year={2023},
  volume={23},
  number={13},
  pages={14471-14486},
  doi={10.1109/JSEN.2023.3277042}}

@inproceedings{stevens2024bioclip,
      title = {{B}io{CLIP}: A Vision Foundation Model for the Tree of Life}, 
      author = {Samuel Stevens and Jiaman Wu and Matthew J Thompson and Elizabeth G Campolongo and Chan Hee Song and David Edward Carlyn and Li Dong and Wasila M Dahdul and Charles Stewart and Tanya Berger-Wolf and Wei-Lun Chao and Yu Su},
      booktitle={Proceedings of the IEEE/CVF Conference on Computer Vision and Pattern Recognition (CVPR)},
      year = {2024},
      pages = {19412-19424}
    }

@misc{ke2025marigold,
  title={Marigold: Affordable Adaptation of Diffusion-Based Image Generators for Image Analysis},
  author={Bingxin Ke and Kevin Qu and Tianfu Wang and Nando Metzger and Shengyu Huang and Bo Li and Anton Obukhov and Konrad Schindler},
  year={2025},
  eprint={2505.09358},
  archivePrefix={arXiv},
  primaryClass={cs.CV}
}

@misc{depth-pro,
title = {Depth Pro: Sharp Monocular Metric Depth in Less Than a Second},
author = {Aleksei Bochkovskii and Amaël Delaunoy and Hugo Germain and Marcel Santos and Yichao Zhou and Stephan R. Richter and Vladlen Koltun},
year = {2024},
URL = {https://arxiv.org/abs/2410.02073}
}

@inproceedings{wang2022medclip,
  title={Medclip: Contrastive learning from unpaired medical images and text},
  author={Wang, Zifeng and Wu, Zhenbang and Agarwal, Dinesh and Sun, Jimeng},
  booktitle={Proceedings of the Conference on Empirical Methods in Natural Language Processing. Conference on Empirical Methods in Natural Language Processing},
  volume={2022},
  pages={3876},
  year={2022}
}

@article{10.1650/CONDOR-18-25.1,
author = {Eben H. Paxton and Megan Laut and John P. Vetter and Steve J. Kendall},
title = {{Research and management priorities for Hawaiian forest birds}},
volume = {120},
journal = {The Condor},
number = {3},
publisher = {American Ornithological Society},
pages = {557 -- 565},
year = {2018},
doi = {10.1650/CONDOR-18-25.1},
URL = {https://doi.org/10.1650/CONDOR-18-25.1}
}

@article{10.2984/73.2.1,
author = {Kathryn N. van Dyk and Kristina L. Paxton and Patrick J. Hart and Eben H. Paxton},
title = {{Seasonality and Prevalence of Pollen Collected from Hawaiian Nectarivorous Birds}},
volume = {73},
journal = {Pacific Science},
number = {2},
publisher = {University of Hawaii Press},
pages = {187 -- 197},
year = {2019},
doi = {10.2984/73.2.1},
URL = {https://doi.org/10.2984/73.2.1}
}

\end{document}